\documentclass{article}

\PassOptionsToPackage{numbers, compress}{natbib}

\usepackage[preprint]{neurips_2025}
 \usepackage{amsmath}
\usepackage{fontawesome5}
\usepackage{enumitem}
\usepackage{wrapfig}
\usepackage[table,xcdraw]{xcolor}
\usepackage{multirow}
\usepackage[utf8]{inputenc} 
\usepackage[T1]{fontenc}    
\usepackage{hyperref}       
\usepackage{url}            
\usepackage{booktabs}       
\usepackage{amsfonts}       
\usepackage{nicefrac}       
\usepackage{microtype}      
\usepackage{xcolor}         
\usepackage[most]{tcolorbox}
\usepackage{fontawesome5}
\usepackage{hyperref}
\usepackage{array} 
\usepackage{listings}
\usepackage{graphicx}  
\usepackage{subcaption} 
\usepackage{float}      
\usepackage{fontawesome5}
\usepackage{booktabs}
\usepackage{graphicx}
\usepackage{colortbl}
\usepackage[table]{xcolor}
\definecolor{lightblue}{RGB}{235,245,255} 
\definecolor{liautoblue}{RGB}{71,111,182} 
\definecolor{textred}{RGB}{128,0,0}
\usepackage{titlesec}
\titleformat{\section}
  {\large\bfseries\color{liautoblue}}{\thesection}{1em}{}
\titleformat{\subsection}
  {\bfseries\color{liautoblue}}{\thesubsection}{1em}{}
  \titleformat{\subsubsection}
  {\bfseries\color{liautoblue}}{\thesubsubsection}{1em}{}
\usepackage[most]{tcolorbox}
\newtcolorbox{liautoabstract}{
    colback=lightblue,
    colframe=white,
    boxrule=0pt,
    arc=2mm,
    left=4mm,
    right=4mm,
    top=5mm,
    bottom=5mm,
    enhanced, 
    before upper={\setlength{\parindent}{0pt}} 
}
\usepackage[most]{tcolorbox} 

\newtcolorbox{stepTitle}[1]{
    enhanced,
    colback=gray!5,    
    colframe=black!50,  
    boxrule=-1pt,
    arc=0mm,            
    left=2mm, right=2mm, top=1mm, bottom=1mm,
    fontupper=\small,  
    title=#1
}

\tcbset{
    stepbox/.style={
        enhanced,
        colback=gray!20,         
        colframe=gray!50,        
        boxrule=0.5pt,
        title={#1},              
        fonttitle=\bfseries,     
        left=2mm, right=2mm, top=1mm, bottom=1mm
    }
}

\newtcolorbox[auto counter]{case}[2][]{ 
    enhanced,
    colback=gray!5,
    colframe=black!70,
    coltitle=white,
    fonttitle=\bfseries\sffamily,
    fontupper=\small,
    arc=1.5mm,
    boxrule=0.5pt,
    title=Case study \thetcbcounter: #2, 
    left=1mm, right=1mm, top=2mm, bottom=2mm,
    label type=case, 
    #1               
}

\newtcolorbox{toolbox}[1]{
    enhanced,                 
    colback=gray!5,           
    colframe=black!70,        
    coltitle=white,           
    fonttitle=\bfseries\sffamily,
    fontupper=\small,
    arc=1.5mm,                
    boxrule=0.5pt,            
    title=#1,                 
    left=1mm, right=1mm, top=2mm, bottom=2mm
}

\usepackage{fancyhdr}
\usepackage{graphicx} 
\hypersetup{
    colorlinks=true,
    urlcolor=liautoblue,
}

\title{Evaluating the Search Agent in a Parallel World}

\author{%
  Jiawei Chen \And Xintian Shen \And Lihao Zheng \And Lifu Mu \And Haoyi Sun \AND Ning Mao \And Hao Ma\thanks{Technique Leader} \And Tao Wei\footnotemark[1] \And Pan Zhou\thanks{Supervisor} \And Kun Zhan\footnotemark[2]  \\
}

\begin{document}

\maketitle

\begin{liautoabstract} 
Integrating web search tools has significantly extended the capability of Large Language Models (LLMs) to address open-world, real-time, and long-tail problems. However, evaluating these Search Agents presents formidable challenges. First, constructing high-quality deep search benchmarks is prohibitively expensive, while unverified synthetic data often suffers from unreliable sources. Second, static benchmarks face dynamic obsolescence: as internet information evolves, complex queries requiring "deep research" often degrade into simple retrieval tasks due to increased popularity (difficulty collapse), and ground truths become outdated due to temporal shifts (fact drift). Third, attribution ambiguity confounds evaluation, as an agent's performance is often dominated by its parametric memory rather than its actual search and reasoning capabilities. Finally, reliance on specific commercial search engines introduces variability that hampers reproducibility.
To address these issues, we propose a novel framework--Mind-ParaWorld\,(MPW) for evaluating Search Agents in a "Parallel World." Specifically, MPW samples real-world entity names to synthesize "future scenarios \& questions" situated beyond the model's knowledge cutoff. A "ParaWorld Law Model" then constructs a set of indivisible Atomic Facts and a unique ground-truth answer for each question. During evaluation, instead of retrieving real-world results, the agent interacts with a ParaWorld Engine Model\,(PEM) that dynamically generates SERP-style evidence\,(news snippets) grounded in these inviolable Atomic Facts. We release MPW-Bench, an interactive benchmark spanning 19 domains with 1,608 instances. Experiments across three evaluation settings show that, while search agents are strong at evidence synthesis given complete information, their performance is limited not only by evidence collection and coverage in unfamiliar search environments, but also by unreliable evidence sufficiency judgment and when-to-stop decisions—bottlenecks. All resource of this work has been released.

\vspace{3mm}
    {\color{liautoblue!30}\rule{\linewidth}{0.5pt}} 
    \vspace{2mm}

    \small 
    \renewcommand{\arraystretch}{1.3} 
\begin{tabular}{@{} l l @{}}
        {\faCalendar*} & \textbf{Last Update Date:} March 4, 2026 \\
        {\color{liautoblue}\faEnvelope} & \textbf{Correspondence:} {chenjiawei13@lixiang.com, zhankun@lixiang.com} \\
        {\faGithub} & \textbf{Code:} \href{https://github.com/TIMMY-CHAN/Mind-ParaWorld}{https://github.com/TIMMY-CHAN/Mind-ParaWorld} \\
        {\raisebox{-0.5ex}{\includegraphics[height=1.1em]{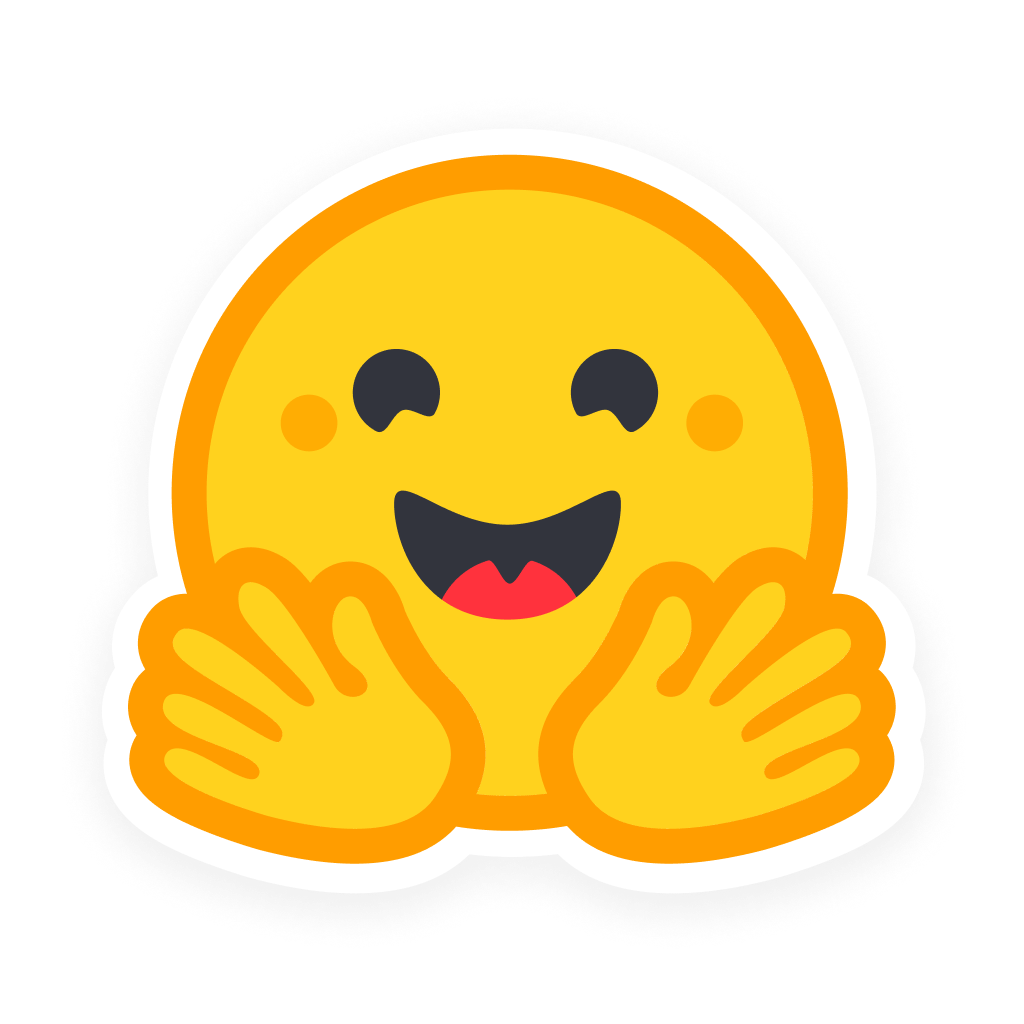}}} & \textbf{Hugging Face:} \href{https://huggingface.co/datasets/LiAutoAISiliconLab/Mind-Paraworld-Bench}{LiAutoAISiliconLab/Mind-Paraworld-Bench}
    \end{tabular}\end{liautoabstract}

\section{Introduction}
Large Language Models (LLMs)~\cite{gpt, GPT-4-REPORT, deepseek-r1, yang2024qwen2.5, gemini2.5, llama,mindgpt-4ov} have revolutionized information processing but remain constrained by two fundamental limitations: temporal staticity, where fixed parameters impose a strict knowledge cutoff, and long-tail deficits, which hinder performance on obscure or specialized queries. While Retrieval-Augmented Generation (RAG)~\cite{RAG} systems and workflow paradigm~\cite{shi2025flowagent, li2025mccd} offer a partial remedy, they are typically limited to single-step retrieval and lack the agency to refine their search scope. In contrast, Search Agents~\cite{ li2025webweaver, wu2025webdancer, chen2025mindwatcher} operating under the ReAct~\cite{yao2022react} paradigm represent a significant leap forward. These agents function as autonomous, iterative systems capable of problem decomposition and strategic planning. By dynamically adjusting their search trajectories based on intermediate feedback, Search Agents can synthesize multi-source information to solve "Deep Search" problems that are intractable via simple, direct queries.

Despite the rapid evolution of Search Agents, evaluation methodologies have failed to keep up with the times. Traditional benchmarks~\cite{infoseek} measuring isolated fact retrieval, such as SimpleQA~\cite{simpleqa}, have become saturated by models equipped with basic browsing capabilities. Consequently, current approaches~\cite{wu2025webwalker} to evaluating "Deep Search" capabilities—primarily relying on static datasets—face three fundamental challenges:

\paragraph{1. Dynamic Obsolescence of Static Benchmarks:} The real-world search environment is highly fluid. As information accumulates on the web, complex problems requiring multi-hop reasoning often degrade into trivial retrieval tasks due to the emergence of direct sources—a phenomenon we term "Difficulty Collapse." Furthermore, "Fact Drift" (e.g., changes in sports records, stock rankings, or pricing) renders the ground truth of static benchmarks rapidly outdated.

\paragraph{2. Attribution Ambiguity:} The distinction between parametric memory and genuine agentic reasoning remains a critical scientific hurdle. Given the vast pre-training corpora of LLMs, it is often unclear whether a model solves a problem through tool usage or merely by reciting memorized long-tail knowledge. When tools fail to provide all necessary clues, evaluation degenerates into a test of knowledge retention rather than search capability.

\paragraph{3. The Cost-Quality Paradox:} Constructing high-quality deep search benchmarks requires prohibitive expert effort, limiting scalability. In contrast, fully automated pipelines often yield synthetic data plagued by logical loopholes or unreliable sources, compromising robustness. Moreover, reliance on commercial search engines introduces uncontrollable engine bias via opaque ranking algorithms, creating an unfair comparison environment.

\begin{figure}
    \centering
    \includegraphics[trim={2cm 3.7cm 3.3cm 2.0cm}, clip, width=\textwidth]{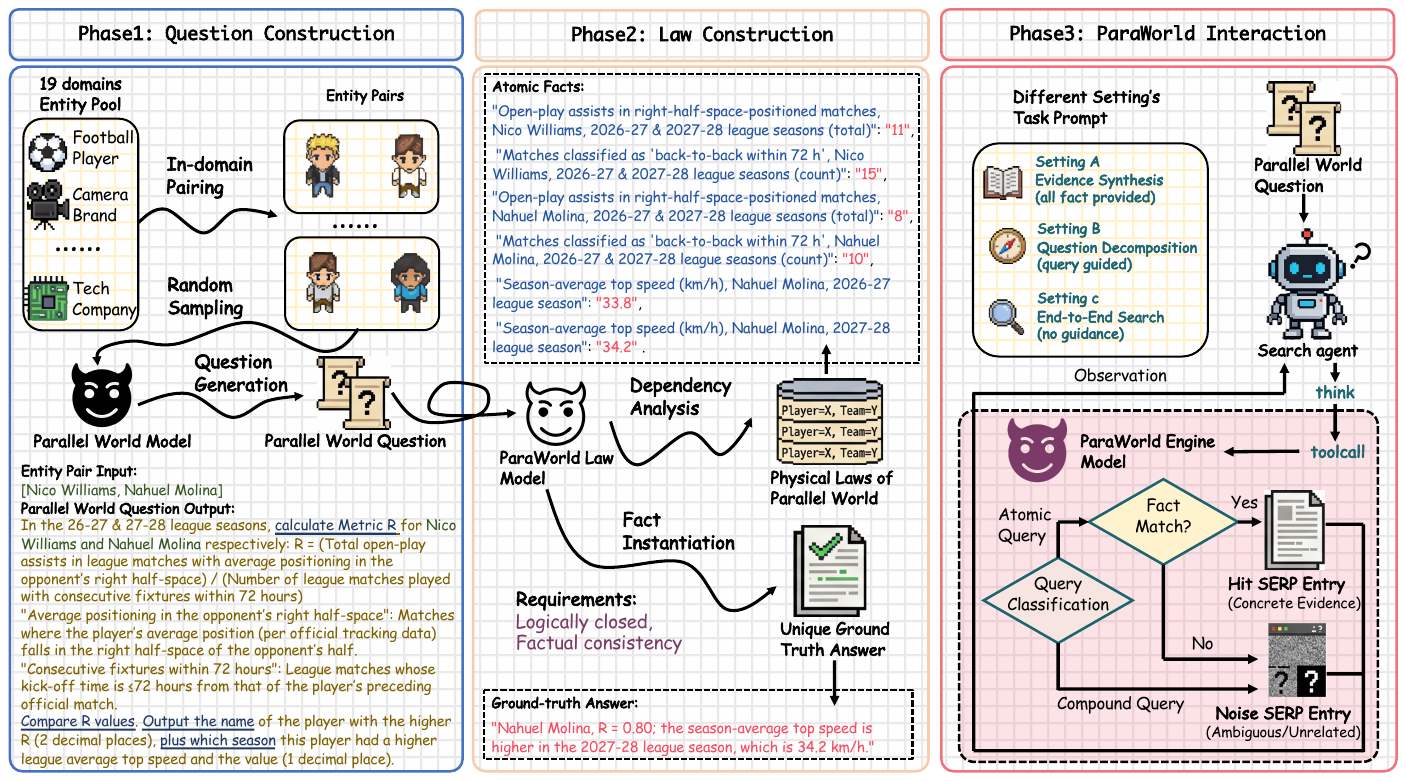}
    \caption{Overview of Mind-ParaWorld Framework.}
    \label{fig:main_arch}
\end{figure}

To address these challenges, we propose Mind-ParaWorld (MPW), an evaluation framework that constructs a dynamic “parallel world” cognitively isolated from the agent’s internal knowledge. Figure~\ref{fig:main_arch} provides an overview of the MPW pipeline. 
Specifically, MPW samples real-world entity names and uses a Parallel World Model to extrapolate future-situated scenarios from entity pairs, formulating them as questions beyond the model’s knowledge cutoff. A ParaWorld Law Model then decomposes each question into a set of inviolable, indivisible Atomic Facts (the “physical laws” of the scenario) and derives a unique ground-truth answer under these laws. This design ensures that retrievable information is logically closed and factually consistent, mitigating the inconsistencies common in uncontrolled synthetic benchmarks.

In MPW, the real-world search engine is replaced by a ParaWorld Engine Model (PEM). Instead of retrieving static web pages, the agent interacts with PEM, which dynamically generates news snippets grounded in the scenario’s Atomic Facts. Crucially, MPW enforces an anti-shortcut mechanism: evidence availability is conditioned on query granularity, suppressing broad, underspecified queries and rewarding decomposed, fine-grained queries with fact-grounded evidence, thereby reducing shortcut-induced confounding and ensuring that the evaluation signal reflects decomposition and evidence-seeking.

Building on the proposed MPW method, we further introduce a novel evaluation benchmark, MPW-Bench. MPW-Bench adopts real-world entity names as construction anchors and curates entity pools spanning 19 domains. Within each domain, entity pairs are deliberately constructed through controlled sampling, systematically injecting compositional diversity while preserving contextual realism and domain coherence. This pairing strategy naturally gives rise to information need patterns prevalent in deep search tasks—including comparison, filtering, aggregative computation, and rule-specification judgment. We then employ the Parallel World Model to generate a series of counterfactual-scenario questions and subject all samples to automated consistency and credibility auditing. The resulting benchmark, MPW-Bench, comprises 1,608 instances spanning multiple levels of complexity.

Our main contributions are as follows:
  \begin{itemize}[leftmargin=*, noitemsep, topsep=2pt]
      \item We propose \textbf{Mind-ParaWorld}, a \textbf{novel parallel-world evaluation paradigm} that constructs a cognitively isolated and controllable search environment for assessing deep-search agents beyond static, real-world web-dependent benchmarks.
      \item We release \textbf{MPW-Bench}, an interactive benchmark grounded in real-world entities, spanning \textbf{19} domains with \textbf{1,608} scenarios of varying complexity, enabling reproducible and fair comparisons across search agents.
      \item We develop a \textbf{decomposed evaluation protocol} with three complementary settings and process-aware diagnostics, enabling \textbf{stage-wise attribution} of end-to-end failures and \textbf{behavioral analysis} beyond final accuracy.
  \end{itemize}
\section{The Mind-ParaWorld Framework}
\subsection{Definition of the Four Stages of Agent Development}
To clarify evaluation objectives, in this paper, we define the capability evolution of Tool Integrated Reasoning\,(TIR) agents as four phases:

\textbf{Phase 1 (Tool Execution Syntax):} The agent knows how to invoke tools with correct syntax.  

\textbf{Phase 2 (Selection):} The agent knows whether to invoke tools for a task, performs simple multi-step tool use, and engages in multi-turn reasoning incorporating tool-call results.  

\textbf{Phase 3 (Decomposition \& Planning):} When facing complex problems, the agent knows how to decompose the problem, perform reasoning and action planning, and translate this understanding into precise sequences of tool calls.  

 \textbf{Phase 4 (Adaptation \& Correction):} When the optimal action and reasoning path is blocked (e.g., the environment does not return expected information), the agent can correct the path based on environmental feedback and find alternative solutions to solve the problem (Resilience  Error Handling).

MPW is specifically designed to evaluate Search Agents equipped with the web search tool at Phase 3 and Phase 4.

\subsection{Overview of Mind-ParaWorld Framework}

The core idea of MPW is to construct a controllable parallel-world environment that is cognitively isolated from a model’s parametric memory, thereby enabling a cleaner evaluation of Search Agents. As illustrated in Figure~\ref{fig:main_arch}, MPW consists of three stages: (1) parallel-world question construction, (2) parallel-world law construction, and (3) agent--environment interaction for evaluation. Specifically, in the first stage, a \emph{Parallel World Model} generates future-situated questions grounded in sampled real-world entities. In the second stage, a \emph{ParaWorld Law Model} decomposes each question into a set of inviolable \emph{Atomic Facts} and derives a unique ground-truth answer under these laws. In the third stage, a \emph{ParaWorld Engine Model} provides a unified web-search interface and returns fact-grounded evidence as environment feedback, through which the agent iteratively collects evidence and produces the final answer. We next describe how MPW constructs
   ParaWorld questions and the underlying principles that guide their design.
\subsection{Construction of the ParaWorld Questions}
\label{sec:paraworld_questions}
MPW evaluates Search Agents primarily at Phase 3 and Phase 4. Its goal is not to test whether an agent can retrieve a single isolated fact, but whether it can, given a complex question, identify the required information needs, issue actionable queries, and iteratively refine its search trajectory through multi-turn interaction to assemble evidence and synthesize a final answer. 
The design of ParaWorld questions follows three principles: multi-fact dependency, parametric-memory isolation, and anti-shortcut.

\paragraph{Multi-fact dependency.}
Deep-search problems in real settings typically involve multiple constraints (e.g., time, location, measurement conventions, target entities, and conditional filters) and require aggregating evidence from multiple sources. Motivated by this observation, MPW designs ParaWorld questions to inherently require multiple facts: each question is associated with an implicit fact-dependency graph, and the agent must identify and verify multiple information nodes via multi-turn search before it can complete the final computation or decision.

\paragraph{Parametric-memory isolation.}
ParaWorld questions place the key facts in a future time range beyond the model’s knowledge cutoff. The purpose is to ensure that, structurally, the answer cannot be derived from prior knowledge alone: without using tools to progressively verify intermediate facts, the model cannot infer the correct answer. In other words, the design targets a “must-search” solution path, rather than merely creating unfamiliarity by shifting timestamps.

\paragraph{Anti-shortcut.}
MPW adopts an anti-shortcut principle: while a question may be complex in surface form, its solvable path should decompose into multiple actionable and locatable search sub-goals. As a result, high-quality queries correspond to the precise retrieval of specific sub-information, rather than issuing the original question as a single monolithic query. This principle ensures that the downstream evaluation results primarily reflect the search agent’s decomposition, planning, and evidence synthesis, rather than whether a search engine happens to return an aggregated SERP containing answers directly.

Based on these principles, we employ a Parallel World Model to construct ParaWorld questions. As shown in Figure~\ref{fig:main_arch}, for a given set of related entity pairs, the model generates multiple questions with increasing levels of difficulty. The difficulty increase is reflected in the structural complexity of the underlying task graph: from cases where sub-goals are explicit, and subproblems are parallel and largely independent, to cases where sub-goals are implicit and involve multi-hop dependencies in which later sub-queries depend on intermediate results; meanwhile, the task objective evolves from a single overall goal to multiple goals and multiple required outputs.

\subsection{Constructions of the ParaWorld Laws}
\label{sec:paraworld_laws}

To ensure that the parallel-world evaluation environment is controllable and reproducible, MPW constructs a set of \emph{ParaWorld Laws} for each ParaWorld question. We define these laws as a collection of inviolable \emph{Atomic Facts} that jointly serve as the “physical laws” of the parallel world. During evaluation, all information synthesized by the environment must be consistent with these laws, and the ground-truth answer to the question must be uniquely determined under them.

\paragraph{Representation of Atomic Facts.}
As shown in Figure~\ref{fig:main_arch}, in MPW, Atomic Facts are stored as key--value pairs in the form of \emph{atomic query description → atomic answer}, where the key is a natural-language description of a minimal information unit, and the value is its unique instantiation in the parallel world (e.g., a number, an entity assignment, a Boolean judgment, or a short text span). We additionally allow a special type of Atomic Fact that specifies \emph{computation or convention rules}. For example, a question may require aggregation under a particular seat-count basis, discount rule, or compliance convention; since such rules are indispensable for solving the task, they are treated as part of the ParaWorld Laws and stored alongside other Atomic Facts.

The construction of ParaWorld Laws follows three core objectives:
(1) \textbf{Consistency:} Atomic Facts must not conflict with one another.
(2) \textbf{Atomicity \& Sufficiency:} each Atomic Fact should correspond to a minimal, independently retrievable information unit (i.e., it cannot be further decomposed into smaller searchable facts). Meanwhile, the fact set should be sufficient to determine the answer, covering all information required for solving the question, including necessary computation or convention rules.
(3) \textbf{Uniqueness:} given the set of Atomic Facts and their implied constraints, the question should admit a unique ground-truth answer, ensuring that evaluation remains well-defined.

Based on these objectives, given a ParaWorld question, we use a \emph{ParaWorld Law Model} to analyze the question, identify the key intermediate variables and information dependencies required for solving it, and instantiate them as a set of Atomic Facts (key--value pairs). This process does not aim to reproduce any single fixed decomposition path; rather, it extracts the lowest-level truth anchors needed to answer the question, such that a Search Agent can progressively hit them through appropriate queries during interaction. After generating the Atomic Facts, the ParaWorld Law Model performs explicit logical reasoning or numerical computation under its own specified rules to derive the final ground-truth answer, and then self-checks the generated facts for coherence and non-decomposability.

\subsection{Generation of the ParaWorld News}
\label{sec:paraworld_news}
During MPW evaluation, we replace the real-world search engine with a \emph{ParaWorld Engine Model} (PEM), which provides a unified \texttt{web\_search} tool interface for Search Agents. Given the original question, the corresponding ParaWorld Laws (i.e., the set of Atomic Facts), and the agent’s current query, PEM generates a simulated SERP, where each entry contains a title, a content snippet, and a timestamp. All information generated by PEM must strictly use the Atomic Facts as the sole source of truth, thereby ensuring that environment feedback is both consistent and controllable.

\paragraph{Query type classification: atomic vs.\ compound queries.}
As shown in Figure~\ref{fig:main_arch}, PEM first performs a structural classification of the query, assigning it to either an \emph{Atomic Query} or a \emph{Compound Query}. Here, “atomic” means that the query is intended to locate \emph{one} minimal information unit (typically anchored on a single entity and a single attribute/relation), whereas compound queries bundle multiple entities/attributes or express comparison/aggregation/analysis intent that cannot be mapped to a single retrievable fact. Importantly, this classification depends only on the query’s structure and intent, and is independent of whether a matching fact exists in the ParaWorld Laws; in particular, an atomic query that misses all facts should not be treated as compound.

If the query is classified as atomic, PEM further checks whether it can be matched to a single Atomic Fact in the ParaWorld Laws: a successful match yields \texttt{hit=1} and returns the corresponding fact key (as \texttt{matched\_fact\_keys}, with at most one element), while a failed match yields \texttt{hit=0}. If the query is classified as compound or low-quality, PEM enforces \texttt{hit=0}. During evaluation, PEM returns a hit log along with the SERP, which is used for downstream process-level metrics and behavioral diagnostics.

PEM always produces a fixed number of search results (four entries), while controlling the amount of actionable information based on availability:
(1) For atomic queries with \texttt{hit=1}, PEM generates four results related to the query and guarantees that at least one “truth entry” explicitly contains the value of the matched Atomic Fact; the remaining entries provide relevant background but must not introduce any facts that contradict the ParaWorld Laws.
(2) For atomic queries with \texttt{hit=0}, PEM still returns four topically related background entries, but must not reveal any concrete values from the Atomic Facts.
(3) For compound/low-quality queries, PEM returns four high-entropy noisy entries consisting of superficially relevant commentary, speculation, or peripheral information, and explicitly avoids leaking any Atomic Fact values or the final answer.

This combination of \emph{query gating} and \emph{controlled information provision} constitutes MPW’s anti-shortcut mechanism: strategies that directly issue the original question as a single query, or overload one query with multiple subgoals, are unlikely to obtain actionable evidence. Only when the agent decomposes the task into locatable atomic queries does the environment return deterministic evidence snippets. As a result, the evaluation signal more directly reflects the agent’s decomposition, search planning, and evidence synthesis capabilities, rather than being confounded by real-world ranking biases or the dynamics of web content.

\section{The Construction of the MPW-Bench}
\subsection{Collection of the Real-world entities and In-domain Entity Pairing and Sampling}
\label{sec:bench_construction}

MPW-Bench is grounded in real-world entities to ensure realistic contexts and broad domain coverage. We curate an entity pool spanning 19 domains\,(as listed in Table~\ref{tab:category_distribution}), including sports, entertainment, consumer products, and technology. This entity pool provides topical boundaries and semantic constraints for downstream ParaWorld question generation, allowing questions to naturally take the form of comparisons, conditional filtering, numerical aggregation, and compliance-style decision making that commonly arise in deep-search scenarios.

To introduce compositional diversity while keeping the domain context fixed, we only form \emph{in-domain} entity pairs by pairing entities within the same domain, and use these pairs as seeds for question construction. Such entity pairing encourages multi-fact, constraint-based information needs (e.g., comparisons under specific conditions) rather than simple single-query fact lookup.

For each entity pair, we aim to generate four questions using the Parallel World Model. We first sample a total of 500 entity pairs. To achieve balanced coverage across the 19 domains, we sample uniformly from each domain-specific entity pool; when a domain contains fewer entities than required for uniform sampling, we compensate by incrementally sampling additional pairs from other domains in a balanced manner. Within each domain, entity pairs are selected via random sampling. 

\subsection{Construction of Questions, Laws and Ground-truth}
\label{sec:scenario_instantiation}

For each sampled entity pair, we generate four ParaWorld questions. For every question, we then construct the corresponding ParaWorld Laws as a set of Atomic Facts (which may include rule/convention facts such as metric definitions or aggregation conventions), and derive a unique ground-truth answer under these laws. Each scenario is thus finalized in the form of \emph{question--laws--answer}, and can be directly used for offline, interactive evaluation.

Applying this pipeline, we obtain a stage-0 version of MPW-Bench consisting of 19 domains with 2{,}000 questions, each paired with its Atomic Facts and ground truth. We then perform automated quality auditing over the entire stage-0 benchmark using Gemini 3 Pro in two steps. In the first step, the auditor checks that (i) the ParaWorld Laws are internally non-contradictory, and (ii) the laws satisfy basic plausibility constraints grounded in real-world commonsense (e.g., realistic travel times and valid units). In the second step, the auditor is provided with the full set of Atomic Facts and asked to answer the question; the resulting answer is compared against the provided ground truth for consistency. We repeat this second-step check three times to reduce stochasticity. This automated auditing filters out 392 instances (19.6\%), yielding the final MPW-Bench with 1{,}608 instances. We assign three \emph{complexity tiers} post-hoc based on the number of atomic facts $|\mathcal{F}|$, and report the resulting distribution in Table~\ref{tab:mpw_complexity_tiers}. The final distribution of categories has been shown in Table~\ref{tab:category_distribution}.

\begin{table*}[]
\centering
\small
\caption{Complexity tiers in MPW-Bench, defined post-hoc by the number of atomic facts $|\mathcal{F}|$ in each instance.}
\begin{tabular}{lccc}
\toprule
\textbf{Tier} & $\mathbf{|\mathcal{F}|}$ \textbf{range} & \textbf{\#Instances} & \textbf{Percentage\%} \\
\midrule
Easy & 1--5  & 568 & 36.52 \\
Mid  & 6--10 & 619 & 38.50 \\
Hard & $\ge$ 11 & 421 & 26.18 \\
\bottomrule
\end{tabular}
\label{tab:mpw_complexity_tiers}
\end{table*}

\begin{table}[]
\centering
\caption{Distribution of categories in the MPW-Bench.}
\label{tab:category_distribution}
\resizebox{\textwidth}{!}{%
\begin{tabular}{lr lr lr lr lr}
\toprule
\textbf{Category} & \textbf{\#} & \textbf{Category} & \textbf{\#} & \textbf{Category} & \textbf{\#} & \textbf{Category} & \textbf{\#} & \textbf{Category} & \textbf{\#} \\
\midrule
\rowcolor{gray!8}
LaLiga Player      & 100 & Serie Player      & 100 & Bundesliga Player & 92 & Ligue1 Player    & 88 & Premier Player    & 88 \\
Football Team      &  96 & F1 Player         &  84 & NBA Player        & 80 & Singer           & 100 & Actor             & 80 \\
\rowcolor{gray!8}
Game Company       &  88 & LOL Player        &  80 & Camera Brand      & 88 & Computer Brand   &  32 & Phone Brand       & 56 \\
Fashion Brand      &  96 & Beauty Brand      &  92 & Car Brand         & 92 & Tech Company     &  76 &                   &   \\
\midrule
\multicolumn{8}{l}{\textbf{Total}} & \multicolumn{2}{r}{\textbf{1608}} \\
\bottomrule
\end{tabular}%
}
\end{table}
\subsection{Data Format and Benchmark Interface}
\label{sec:bench_interface}

MPW-Bench is constructed as a collection of interactive \emph{scenarios}. Each scenario instantiates a ParaWorld question together with its ParaWorld Laws (Atomic Facts) and a unique ground-truth answer, enabling offline evaluation under a controlled search environment.

\paragraph{Data representation.}
Each instance contains (i) a future-situated question, (ii) a set of Atomic Facts expressed as natural-language key--value pairs (including optional rule/convention facts such as metric definitions), and (iii) the derived ground-truth answer. The data sample is shown in the appendix.

\paragraph{Interactive protocol.}
Evaluation is conducted with a single tool, \texttt{web\_search}. A PEM implements this tool by producing a simulated SERP for each agent query, strictly grounded in the Atomic Facts of the current scenario. In addition to the SERP, PEM outputs a hit log indicating whether the query is considered compound/low-quality and whether it matches an atomic fact, enabling process-level analysis of agent behaviors such as fact coverage and query efficiency without altering the agent’s interaction interface. In our implementation, PEM returns a fixed-size SERP with four entries per query and logs at most one matched atomic-fact key for each tool call.

\section{Experiments}
\subsection{Metrics for MPW-Bench}

In MPW, we primarily evaluate search agents on MPW-Bench using \textbf{pass@1}. In addition, we introduce several process-aware metrics to characterize an agent’s overall performance.

\paragraph{Fact Coverage Rate (FCR).}
To measure whether an agent retrieves sufficient evidence, we define the \textbf{Fact Coverage Rate}. For each instance, let $\mathcal{F}$ denote the set of atomic fact keys in the scenario, and let $\mathcal{H}$ denote the set of unique fact keys hit by the agent across all tool calls (obtained by deduplicating matched fact). We compute:
\[
\mathrm{FCR}=\frac{|\mathcal{H}|}{|\mathcal{F}|}.
\]

\paragraph{Hit Rate.}
We further quantify the fraction of an agent’s queries that effectively hit an atomic fact. Let $T$ be the total number of tool calls for an instance, and let $H$ be the number of hits (i.e., calls with \texttt{hit=1}). We define:
\[
\mathrm{HitRate}=\frac{H}{T}.
\]

\subsection{Application Details}
\label{sec:app_details}
To better interpret the performance of Search Agents on MPW-Bench, we design three evaluation settings of increasing difficulty, which respectively characterize (i) an upper bound of evidence synthesis, (ii) decomposition and evidence coverage under guided querying, and (iii) fully end-to-end capability spanning decomposition, query strategy, coverage, and synthesis. For the agent mode, we evaluate all models using the ReAct paradigm.

Specifically, \textbf{Setting A (Oracle-Facts QA, upper bound)} directly provides the model with the original question and the complete set of atomic facts for that instance (including necessary computation/convention rules), and forbids any tool calls. This setting evaluates the model’s ability to integrate information and synthesize the final answer given all required evidence and conditions, serving as an achievable upper bound once all necessary information has been obtained through search.

\textbf{Setting B (Guided Search)} retains the end-to-end interactive form: the model still needs to decompose the problem and interact with PEM via \texttt{web\_search} to collect evidence, but we explicitly guide the model through the system prompt to construct “hittable” atomic queries (e.g., single entity + single attribute), so as to reduce confounding from query formulation quality. To differentiate between two guidance strengths—lightweight rule prompting versus stronger decomposition exemplars—Setting B includes two prompt variants: the \emph{Guidance Prompt} provides a concise guide for query construction and query decomposition examples, while the \emph{Few-shot Decomposition Prompt} not only includes multiple query decomposition examples, but also provides planning processes for solving questions from multiple cases as reference.

\textbf{Setting C (End-to-End Search)} provides only the original question and basic tool instructions, without any additional query guidance. The model must independently decompose the problem, formulate queries, infer how to construct high-quality queries from interaction, retrieve sufficient evidence coverage, and synthesize the final answer.
More detailed explanations and the exact prompts for Settings A/B/C are provided in the appendix.

To ensure fair comparison, we apply the same termination conditions to all models: inference stops immediately when the agent outputs \texttt{<answer>...\ </answer>} or when the maximum interaction budget of 32 turns is reached. In our experiments, we use Qwen3-235B-A22B-Thinking as the PEM. For correctness evaluation, we adopt an LLM-as-Judge protocol, the judge model is also Qwen3-235B-A22B-Thinking and the system prompt is provided in the appendix.
\subsection{Setting A Results: Evidence-synthesis upper bound}

Table~\ref{tab:setting_a} reports results under \textbf{Setting A (Oracle-Facts QA)}. Setting A estimates an upper bound on performance once all necessary evidence has been acquired, serving both to validate task solvability and to characterize evidence synthesis and aggregation ability under complete information.

Most models achieve strong Pass@1 under Setting A (67.23\%--91.04\%). Qwen3-32B attains the best overall performance (91.04\% overall; 94.86\% on Easy; 85.75\% on Hard), while GLM-4.7 is also competitive (89.80\% overall). MindWatcher 32B, Qwen3-30B-A3B, and Tongyi DeepResearch all exceed 83\% overall. These results indicate that, given complete evidence, modern models are generally capable of integrating the provided information and synthesizing correct answers, supporting that MPW-Bench is well-posed under atomic-fact constraints.

We further observe performance degradation with increasing complexity for most models (e.g., Qwen3-32B drops from 94.86\% in Easy to 85.75\% in Hard), suggesting that even without retrieval, harder instances require non-trivial multi-condition aggregation, correct application of conventions/rules, and longer-chain reasoning. Importantly, the high accuracy in Setting A provides a clear reference point for the end-to-end evaluations in Settings B/C: when performance in these settings falls far below the Setting A upper bound, the dominant bottleneck is more plausibly evidence acquisition and coverage through search, rather than an inability to reason given complete evidence.
\begin{table}[t]
\centering
\caption{Performance of all models under Setting A (Pass@1 \%). 
\textbf{E}: Easy, \textbf{M}: Medium, \textbf{H}: Hard, \textbf{O}: Overall.}
\label{tab:setting_a}
\resizebox{0.8\textwidth}{!}{%
\renewcommand{\arraystretch}{1.15}
\fontsize{4.5pt}{5.5pt}\selectfont
\begin{tabular}{@{}lcccc@{}}
\toprule
\textbf{Model} & \textbf{Overall} & \textbf{Easy} & \textbf{Medium} & \textbf{Hard} \\
\midrule
MindWatcher 32B~\cite{chen2025mindwatcher}                                                          & 85.01 & 88.83 & 85.23 & 79.57 \\
Qwen3-30B-A3B~\cite{yang2025qwen3}                                                           & 85.14 & 87.23 & 87.16 & 79.33 \\
Qwen3-32B~\cite{yang2025qwen3}                                                               & \textbf{91.04} & \textbf{94.86} & \textbf{91.17} & \textbf{85.75} \\
MiniMax-m2                                                              & 67.23 & 66.67 & 67.58 & 67.46 \\
MiniMax-m2.1                                                            & 82.46 & 84.93 & 79.94 & 82.90 \\
Tongyi DeepResearch~\cite{tongyideep} & 83.40 & 83.33 & 83.15 & 83.85 \\
Youtu Agent 7B~\cite{shi2025youtu}                                                          & 67.91 & 78.90 & 67.42 & 53.92 \\
GLM-4.7                                                      & 89.80 & 92.91 & 90.69 & 84.32 \\
\bottomrule
\end{tabular}
}
\end{table}

\subsection{Setting B Results: Decomposition and Coverage under Query Guidance}
Table~\ref{tab:main_results} summarizes results under Setting B. Pass@1, FCR, and Hit Rate are reported in percentages (\%), and ToolCalls denotes the number of search queries issued by the model.

\textbf{Overall performance and model differences.}
On the Overall split, MindWatcher 32B achieves the best performance under both prompts (47.51\% with the Fewshot prompt and 44.15\% with the Guidance prompt), together with high FCR and Hit Rate (both in the 40--50\% range), indicating that it can stably cover key facts through multi-turn querying and complete final synthesis under guided search. MiniMax-m2.1 and MiniMax-m2 also perform strongly, reaching 37.19\%/39.99\% and 32.09\%/34.33\% Pass@1, respectively, suggesting relatively strong evidence coverage and aggregation. Qwen3-32B reaches 27.24\% under the Guidance prompt, and Tongyi DeepResearch reaches 19.09\%. In contrast, Youtu Agent 7B remains substantially weaker (5.47\% with Fewshot and 4.23\% with Guidance) with near-zero FCR, indicating difficulty in forming a stable ``retrieval coverage--evidence synthesis'' loop under this setting.

\textbf{Fewshot vs.\ Guidance.}
The gains from the two system prompts are not uniform across models, exhibiting clear model dependence. For example, MindWatcher 32B performs better with the Fewshot prompt than with the Guidance prompt (47.51\% vs.\ 44.15\%), whereas Qwen3-32B, MiniMax-m2/m2.1, and Tongyi DeepResearch are generally stronger or more stable under the Guidance prompt. Overall, given that both prompts already specify query-construction rules, adding a small number of examples and planning references can be beneficial but does not provide a consistent advantage across all models, suggesting that ``rule-based constraints'' and ``example-driven guidance'' elicit different search behaviors depending on the underlying model.

\textbf{Complexity scaling and the coverage bottleneck.}
Across difficulty tiers, Setting B performance degrades noticeably as complexity increases, in tandem with declines in FCR and Hit Rate. For instance, MindWatcher 32B (Guidance) drops from 59.93\% on Easy to 44.14\% on Medium and further to 23.04\% on Hard; correspondingly, FCR decreases from 56.40\% to 36.50\% and 21.80\%, and Hit Rate decreases from 59.30\% to 45.90\% and 35.40\%. Similar trends hold for other models: as task complexity grows, evidence coverage and query-hit quality decrease substantially. Meanwhile, stronger agents such as MindWatcher and MiniMax-m2/m2.1 typically increase their ToolCalls with difficulty, indicating that they adjust search budget in response to task complexity. Nevertheless, increased tool usage does not fully offset the drop in coverage. This provides additional evidence that MPW-Bench's complexity tiers reliably scale task difficulty and place increasing demands on search planning and evidence coverage.

\textbf{Tool usage and stopping behavior.}
ToolCalls further reveal substantial differences in interaction strategies under guided search. MindWatcher 32B typically issues more queries (about 10 on Overall and over 14 on Hard), accompanied by higher Hit Rate and FCR. In contrast, Qwen3-32B achieves relatively competitive accuracy with far fewer tool calls (about 2.6 on Overall), suggesting higher query efficiency. However, a broader pattern emerges: many models output a final answer before achieving sufficient evidence coverage, exhibiting a clear tendency of \emph{premature stopping}, even though the system prompt explicitly provides hittable query-construction rules. This is especially pronounced for weaker models: for example, ToolCalls for Qwen3-30B-A3B and Qwen3-32B change little across difficulty tiers, implying limited expansion of search scope as complexity increases. Moreover, even for models that substantially increase ToolCalls (e.g., MiniMax and MindWatcher), FCR still decreases steadily with difficulty, indicating that simply making more queries is insufficient to guarantee coverage of all key facts. Overall, these results suggest that most current open-source models still lack a reliable decision mechanism for when to continue searching versus when to stop, and tend to answer under insufficient evidence, which limits end-to-end performance on high-complexity tasks.

\begin{table*}[]
\centering
\caption{Performance comparison of all models under Setting B\,(Fewshot and Guidance prompts) across difficulty levels. 
\textbf{E}: Easy, \textbf{M}: Medium, \textbf{H}: Hard, \textbf{O}: Overall. 
Pass@1, FCR, and Hit Rate are reported in percentages (\%).}
\label{tab:main_results}
\resizebox{\textwidth}{!}{%
\renewcommand{\arraystretch}{1.15}
\fontsize{8.5pt}{10.5pt}\selectfont
\begin{tabular}{@{}cl|cccc|cccc@{}}
\toprule
\multirow{2}{*}{\textbf{Model}} & \multirow{2}{*}{\textbf{Diff.}} 
& \multicolumn{4}{c|}{\textbf{Fewshot Prompt}} 
& \multicolumn{4}{c}{\textbf{Guidance Prompt}} \\ 
\cmidrule(l){3-10}
& & \textbf{Pass@1} & \textbf{FCR} & \textbf{HitRate} & \textbf{ToolCalls} 
  & \textbf{Pass@1} & \textbf{FCR} & \textbf{HitRate} & \textbf{ToolCalls} \\ 
\midrule
& E & 59.75 & 56.20 & 58.00 & 6.95  & 59.93 & 56.40 & 59.30 & 6.18 \\
& M & 47.67 & 38.80 & 48.00 & 10.22 & 44.14 & 36.50 & 45.90 & 10.43 \\
& H & 30.88 & 25.30 & 40.00 & 14.15 & 23.04 & 21.80 & 35.40 & 14.53 \\
\multirow{-4}{*}{MindWatcher 32B}
& \cellcolor[HTML]{F8E7E3}O 
& \cellcolor[HTML]{F8E7E3}\textbf{47.51} & \cellcolor[HTML]{F8E7E3}41.40 
& \cellcolor[HTML]{F8E7E3}49.40 & \cellcolor[HTML]{F8E7E3}10.10 
& \cellcolor[HTML]{F8E7E3}\textbf{44.15} & \cellcolor[HTML]{F8E7E3}39.60 
& \cellcolor[HTML]{F8E7E3}47.80 & \cellcolor[HTML]{F8E7E3}10.01 \\
\midrule
& E & 34.04 & 29.50 & 41.10 & 1.98 & 24.12 & 20.20 & 27.00 & 1.89 \\
& M & 17.50 & 13.20 & 33.00 & 2.22 & 13.89 &  9.70 & 22.60 & 2.00 \\
& H &  8.55 &  5.90 & 23.80 & 2.27 &  7.36 &  4.00 & 19.10 & 1.98 \\
\multirow{-4}{*}{Qwen3-30B-A3B}
& \cellcolor[HTML]{F8E7E3}O 
& \cellcolor[HTML]{F8E7E3}\textbf{20.96} & \cellcolor[HTML]{F8E7E3}17.00 
& \cellcolor[HTML]{F8E7E3}33.40 & \cellcolor[HTML]{F8E7E3}2.15 
& \cellcolor[HTML]{F8E7E3}\textbf{15.80} & \cellcolor[HTML]{F8E7E3}11.90 
& \cellcolor[HTML]{F8E7E3}23.20 & \cellcolor[HTML]{F8E7E3}1.96 \\
\midrule
& E & 37.77 & 37.80 & 48.90 & 2.35 & 41.55 & 40.00 & 53.20 & 2.27 \\
& M & 20.39 & 17.10 & 33.90 & 2.82 & 23.10 & 19.60 & 40.90 & 2.67 \\
& H & 10.21 &  6.50 & 22.20 & 2.80 & 14.01 &  8.70 & 30.80 & 2.87 \\
\multirow{-4}{*}{Qwen3-32B}
& \cellcolor[HTML]{F8E7E3}O 
& \cellcolor[HTML]{F8E7E3}\textbf{23.82} & \cellcolor[HTML]{F8E7E3}21.60 
& \cellcolor[HTML]{F8E7E3}36.10 & \cellcolor[HTML]{F8E7E3}2.65 
& \cellcolor[HTML]{F8E7E3}\textbf{27.24} & \cellcolor[HTML]{F8E7E3}24.00 
& \cellcolor[HTML]{F8E7E3}42.60 & \cellcolor[HTML]{F8E7E3}2.58 \\
\midrule
& E & 42.73 & 40.80 & 38.60 & 5.20  & 45.39 & 42.50 & 40.90 & 5.12 \\
& M & 32.42 & 24.10 & 27.30 & 7.24  & 33.23 & 26.90 & 29.80 & 7.78 \\
& H & 17.34 & 13.30 & 18.30 & 9.12  & 21.14 & 14.80 & 20.30 & 10.09 \\
\multirow{-4}{*}{MiniMax-m2}
& \cellcolor[HTML]{F8E7E3}O 
& \cellcolor[HTML]{F8E7E3}\textbf{32.09} & \cellcolor[HTML]{F8E7E3}27.10 
& \cellcolor[HTML]{F8E7E3}28.90 & \cellcolor[HTML]{F8E7E3}7.01 
& \cellcolor[HTML]{F8E7E3}\textbf{34.33} & \cellcolor[HTML]{F8E7E3}29.20 
& \cellcolor[HTML]{F8E7E3}31.20 & \cellcolor[HTML]{F8E7E3}7.45 \\
\midrule
& E & 50.53 & 47.50 & 38.10 & 8.73  & 52.66 & 47.90 & 40.30 & 8.45 \\
& M & 36.28 & 34.70 & 28.30 & 13.13 & 40.45 & 35.60 & 29.20 & 12.97 \\
& H & 20.67 & 22.40 & 21.20 & 16.83 & 22.33 & 20.60 & 22.10 & 15.47 \\
\multirow{-4}{*}{MiniMax-m2.1}
& \cellcolor[HTML]{F8E7E3}O 
& \cellcolor[HTML]{F8E7E3}\textbf{37.19} & \cellcolor[HTML]{F8E7E3}36.00 
& \cellcolor[HTML]{F8E7E3}29.90 & \cellcolor[HTML]{F8E7E3}12.55 
& \cellcolor[HTML]{F8E7E3}\textbf{39.99} & \cellcolor[HTML]{F8E7E3}36.00 
& \cellcolor[HTML]{F8E7E3}31.30 & \cellcolor[HTML]{F8E7E3}12.04 \\
\midrule
& E & 26.60 & 29.80 & 30.80 & 4.72 & 27.99 & 34.70 & 33.90 & 4.66 \\
& M & 15.41 & 16.40 & 25.00 & 5.47 & 17.45 & 21.60 & 29.80 & 5.47 \\
& H &  6.41 &  8.10 & 21.20 & 5.91 &  9.50 & 10.00 & 24.10 & 5.72 \\
\multirow{-4}{*}{\begin{tabular}[c]{@{}c@{}}Tongyi DeepResearch\\[-1pt] 30B-A3B\end{tabular}}
& \cellcolor[HTML]{F8E7E3}O 
& \cellcolor[HTML]{F8E7E3}\textbf{16.98} & \cellcolor[HTML]{F8E7E3}19.00 
& \cellcolor[HTML]{F8E7E3}26.00 & \cellcolor[HTML]{F8E7E3}5.32 
& \cellcolor[HTML]{F8E7E3}\textbf{19.09} & \cellcolor[HTML]{F8E7E3}23.20 
& \cellcolor[HTML]{F8E7E3}29.70 & \cellcolor[HTML]{F8E7E3}5.25 \\
\midrule
& E & 7.62 & 0.30 & 25.00 & 0.04 & 6.21 & 0.10 & 15.40 & 0.03 \\
& M & 4.82 & 0.00 & 13.30 & 0.02 & 3.69 & 0.00 &  8.30 & 0.02 \\
& H & 3.56 & 0.00 &  0.00 & 0.01 & 2.38 & 0.00 &  7.10 & 0.04 \\
\multirow{-4}{*}{Youtu Agent 7B}
& \cellcolor[HTML]{F8E7E3}O 
& \cellcolor[HTML]{F8E7E3}\textbf{5.47}  & \cellcolor[HTML]{F8E7E3}0.10 
& \cellcolor[HTML]{F8E7E3}13.90 & \cellcolor[HTML]{F8E7E3}0.02 
& \cellcolor[HTML]{F8E7E3}\textbf{4.23}  & \cellcolor[HTML]{F8E7E3}0.10 
& \cellcolor[HTML]{F8E7E3}10.50 & \cellcolor[HTML]{F8E7E3}0.03 \\
\bottomrule
\end{tabular}%
}
\end{table*}

\subsection{Setting C Results: End-to-End ReAct}
Tables~\ref{tab:setting_c_local} and~\ref{tab:setting_c_api} report Setting C results for (i) locally deployed models that are aligned with the model set in Settings A/B, and (ii) additional commercial API models included as supplementary evaluation.

\textbf{Overall performance of search agents in an unguided setting.}
From the Overall results in Table~\ref{tab:setting_c_local}, MindWatcher 32B achieves the best performance in this end-to-end setting (Pass@1 38.56\%, FCR 34.90\%, Hit Rate 43.00\%, ToolCalls 8.72), followed by MiniMax-m2.1 (Pass@1 31.84\%, FCR 28.30\%) and MiniMax-m2 (Pass@1 25.44\%, FCR 21.90\%). In contrast, Qwen3-32B and Qwen3-30B-A3B use substantially fewer tool calls (Overall $\sim$1.92 and $\sim$1.59), with correspondingly low FCR (12.30\% and 9.70\%), while Tongyi DeepResearch and Youtu Agent 7B remain lower on both FCR/Hit Rate and Pass@1. Overall, as complexity increases from Easy to Hard, both Pass@1 and FCR drop markedly across models. This systematic degradation indicates that, under unguided end-to-end interaction, models struggle to maintain sufficient evidence coverage on higher-complexity tasks, which in turn limits final correctness.

Table~\ref{tab:setting_c_api} further presents results for commercial API models under Setting C. GPT-5 performs strongly in overall accuracy (Overall Pass@1 33.21\%), but its FCR decreases substantially with difficulty (from 30.80\% on Easy to 8.40\% on Hard), suggesting that even stronger models can be constrained by evidence coverage on harder instances. The Gemini family exhibits more complex interaction patterns. For example, Gemini-3-Pro issues a moderate number of tool calls (Overall 4.35) yet attains low FCR and Hit Rate (6.40\% and 12.70\%), indicating that tool use does not necessarily translate into better coverage and that models may still repeatedly issue low-yield queries. In contrast, Gemini-2.5-Flash makes very few tool calls (0.90) and also has low FCR (6.00\%), which is more consistent with shallow retrieval and early answering. Kimi-K2 reaches an Overall Pass@1 of 20.71\%, but its Hard-bucket FCR is only 1.70\%, again highlighting coverage shortfall as task complexity increases.

\begin{table*}[t]
\centering
\caption{Performance of locally-deployed models under Setting C (End-to-End ReAct).}
\label{tab:setting_c_local}
\resizebox{0.75\textwidth}{!}{%
\renewcommand{\arraystretch}{1.15}
\fontsize{7.5pt}{9.5pt}\selectfont
\begin{tabular}{@{}cccccc@{}}
\toprule
\textbf{Model} & \textbf{Difficulty} & \textbf{Pass@1} & \textbf{FCR} & \textbf{HitRate} & \textbf{ToolCalls} \\
\midrule
& E & 56.69 & 51.30 & 55.30 & 5.78 \\
& M & 36.03 & 31.10 & 40.30 & 9.19 \\
& H & 17.81 & 18.20 & 30.30 & 12.00 \\
\multirow{-4}{*}{MindWatcher 32B}
& \cellcolor[HTML]{F8E7E3}O
& \cellcolor[HTML]{F8E7E3}\textbf{38.56} & \cellcolor[HTML]{F8E7E3}34.90
& \cellcolor[HTML]{F8E7E3}43.00 & \cellcolor[HTML]{F8E7E3}8.72 \\
\midrule
& E & 24.65 & 19.10 & 28.50 & 1.57 \\
& M & 12.60 &  6.10 & 19.00 & 1.61 \\
& H &  6.89 &  2.30 & 14.30 & 1.58 \\
\multirow{-4}{*}{Qwen3-30B-A3B}
& \cellcolor[HTML]{F8E7E3}O
& \cellcolor[HTML]{F8E7E3}\textbf{15.36} & \cellcolor[HTML]{F8E7E3}9.70
& \cellcolor[HTML]{F8E7E3}21.10 & \cellcolor[HTML]{F8E7E3}1.59 \\
\midrule
& E & 28.87 & 24.40 & 34.10 & 1.88 \\
& M & 14.54 &  8.00 & 22.00 & 1.96 \\
& H &  9.74 &  2.20 & 12.40 & 1.90 \\
\multirow{-4}{*}{Qwen3-32B}
& \cellcolor[HTML]{F8E7E3}O
& \cellcolor[HTML]{F8E7E3}\textbf{18.35} & \cellcolor[HTML]{F8E7E3}12.30
& \cellcolor[HTML]{F8E7E3}23.80 & \cellcolor[HTML]{F8E7E3}1.92 \\
\midrule
& E & 38.91 & 35.20 & 35.20 & 3.95 \\
& M & 22.46 & 18.60 & 23.00 & 5.19 \\
& H & 11.64 &  8.90 & 15.30 & 5.76 \\
\multirow{-4}{*}{MiniMax-m2}
& \cellcolor[HTML]{F8E7E3}O
& \cellcolor[HTML]{F8E7E3}\textbf{25.44} & \cellcolor[HTML]{F8E7E3}21.90
& \cellcolor[HTML]{F8E7E3}25.30 & \cellcolor[HTML]{F8E7E3}4.90 \\
\midrule
& E & 47.34 & 42.00 & 33.80 & 7.79 \\
& M & 29.21 & 25.00 & 21.80 & 11.61 \\
& H & 14.96 & 14.60 & 15.10 & 14.08 \\
\multirow{-4}{*}{MiniMax-m2.1}
& \cellcolor[HTML]{F8E7E3}O
& \cellcolor[HTML]{F8E7E3}\textbf{31.84} & \cellcolor[HTML]{F8E7E3}28.30
& \cellcolor[HTML]{F8E7E3}24.30 & \cellcolor[HTML]{F8E7E3}10.92 \\
\midrule
& E & 11.62 &  9.80 & 13.40 & 3.56 \\
& M &  5.82 &  3.80 &  7.10 & 4.11 \\
& H &  4.04 &  0.90 &  3.80 & 4.17 \\
\multirow{-4}{*}{\begin{tabular}[c]{@{}c@{}}Tongyi DeepResearch\\[-1pt] 30B-A3B\end{tabular}}
& \cellcolor[HTML]{F8E7E3}O
& \cellcolor[HTML]{F8E7E3}\textbf{7.40} & \cellcolor[HTML]{F8E7E3}5.10
& \cellcolor[HTML]{F8E7E3}8.50 & \cellcolor[HTML]{F8E7E3}3.93 \\
\midrule
& E &  8.51 &  1.50 &  4.70 & 0.76 \\
& M &  4.33 &  0.20 &  1.90 & 0.73 \\
& H &  3.33 &  0.20 &  2.60 & 0.74 \\
\multirow{-4}{*}{Youtu Agent 7B}
& \cellcolor[HTML]{F8E7E3}O
& \cellcolor[HTML]{F8E7E3}\textbf{5.53} & \cellcolor[HTML]{F8E7E3}0.70
& \cellcolor[HTML]{F8E7E3}3.00 & \cellcolor[HTML]{F8E7E3}0.74 \\
\bottomrule
\end{tabular}%
}
\end{table*}

\vspace{6pt}

\begin{table*}[t]
\centering
\caption{Performance of commercial API models under Setting C (End-to-End ReAct). }
\label{tab:setting_c_api}
\resizebox{0.7\textwidth}{!}{%
\renewcommand{\arraystretch}{1.15}
\fontsize{7.5pt}{9.5pt}\selectfont
\begin{tabular}{@{}clcccc@{}}
\toprule
\textbf{Model} & \textbf{Diff.} & \textbf{Pass@1} & \textbf{FCR} & \textbf{HitRate} & \textbf{ToolCalls} \\
\midrule
& E & 44.19 & 30.80 & 16.60 & 6.69 \\
& M & 30.69 & 16.00 & 10.30 & 9.57 \\
& H & 22.09 &  8.40 &  8.30 & 9.91 \\
\multirow{-4}{*}{GPT-5}
& \cellcolor[HTML]{F8E7E3}O
& \cellcolor[HTML]{F8E7E3}\textbf{33.21} & \cellcolor[HTML]{F8E7E3}19.20
& \cellcolor[HTML]{F8E7E3}12.00 & \cellcolor[HTML]{F8E7E3}8.64 \\
\midrule
& E & 13.73 & 11.20 & 43.30 & 0.75 \\
& M &  5.01 &  4.20 & 30.70 & 0.86 \\
& H &  3.09 &  1.80 & 20.40 & 1.16 \\
\multirow{-4}{*}{Gemini-2.5-Flash~\cite{gemini2.5}}
& \cellcolor[HTML]{F8E7E3}O
& \cellcolor[HTML]{F8E7E3}\textbf{7.59} & \cellcolor[HTML]{F8E7E3}6.00
& \cellcolor[HTML]{F8E7E3}32.40 & \cellcolor[HTML]{F8E7E3}0.90 \\
\midrule
& E &  7.22 &  4.00 & 16.20 & 1.57 \\
& M &  3.55 &  2.10 & 17.00 & 1.37 \\
& H &  4.04 &  0.60 & 11.30 & 1.11 \\
\multirow{-4}{*}{Gemini-2.5-Pro~\cite{gemini2.5}}
& \cellcolor[HTML]{F8E7E3}O
& \cellcolor[HTML]{F8E7E3}\textbf{4.98} & \cellcolor[HTML]{F8E7E3}2.30
& \cellcolor[HTML]{F8E7E3}15.20 & \cellcolor[HTML]{F8E7E3}1.37 \\
\midrule
& E & 27.29 & 11.40 & 14.80 & 4.41 \\
& M & 15.99 &  4.60 & 11.90 & 4.87 \\
& H &  9.98 &  2.20 & 11.10 & 3.50 \\
\multirow{-4}{*}{Gemini-3-Pro}
& \cellcolor[HTML]{F8E7E3}O
& \cellcolor[HTML]{F8E7E3}\textbf{18.41} & \cellcolor[HTML]{F8E7E3}6.40
& \cellcolor[HTML]{F8E7E3}12.70 & \cellcolor[HTML]{F8E7E3}4.35 \\
\midrule
& E & 31.69 & 22.90 & 28.70 & 2.72 \\
& M & 17.45 &  6.80 & 11.80 & 3.26 \\
& H & 10.69 &  1.70 &  4.80 & 3.59 \\
\multirow{-4}{*}{Kimi-K2~\cite{team2025kimik2}}
& \cellcolor[HTML]{F8E7E3}O
& \cellcolor[HTML]{F8E7E3}\textbf{20.71} & \cellcolor[HTML]{F8E7E3}11.10
& \cellcolor[HTML]{F8E7E3}15.90 & \cellcolor[HTML]{F8E7E3}3.16 \\
\bottomrule
\end{tabular}%
}
\end{table*}

To further investigate the termination behavior of mainstream search agents and their interaction logic in unfamiliar search environments, we conduct a deeper analysis under Setting C on the relationship between the number of queries (ToolCalls) and factual coverage. We select three representative models (GPT-5, MindWatcher, and MiniMax-m2.1) for process-level diagnosis. Specifically, for each model we visualize relatively long-horizon trajectories. In the upper plots, the left axis reports the mean number of newly covered atomic facts contributed by the $k$-th tool call, $\Delta \mathrm{facts}(k)$. Samples with $\mathrm{ToolCalls} \ge k_{\text{cohort}}$ constitute the cohort, and the right axis $n(k)$ denotes the number of samples that reach the $k$-th tool call. When $n(k) < 50$, the corresponding region is rendered with reduced opacity, as statistics based on such small sample sizes are typically not representative.

\textbf{Diminishing marginal coverage gains.}
GPT-5 and MindWatcher use the same $k_{\text{cohort}}$. As shown in their cohorts, the mean newly covered atomic facts over the first eight tool calls drops rapidly in the first three calls and then fluctuates thereafter, suggesting that models typically surface high-priority evidence---i.e., salient facts that are comparatively easy to retrieve and crucial for solving the problem---very early in the interaction. For MindWatcher, the marginal gains weaken after $k=12$ and become nearly zero for $k \ge 16$ (while still supported by $n(k)\approx 50+$ samples), indicating that subsequent retrieval increasingly enters a low-yield regime: even with continued tool use, the model rarely acquires additional factual coverage. This effect is further amplified by degradation in long-horizon reasoning, leading to faster decay of marginal gains. In contrast, GPT-5 exhibits sparser long trajectories in the tail, and $n(k)$ decreases rapidly as $k$ grows. This pattern suggests that both MindWatcher and GPT-5 tend to terminate early and answer before collecting sufficient evidence. MiniMax-m2.1, on the other hand, demonstrates stronger evidence awareness: although its early-stage mean marginal gains are smaller than those of GPT-5 and MindWatcher, more than $250$ samples reach $24$ queries; moreover, $\Delta \mathrm{facts}(k)$ does not exhibit a sharp decline with increasing $k$, reflecting better long-horizon reasoning capability.

Taken together, the upper plots show that as $k$ increases, the coverage increment contributed by each additional tool call decreases across all three models, including the model with stronger long-horizon behavior. To further probe query strategies, the lower plots examine the relationship between ToolCalls and (i) cumulative factual coverage $FCR(k)$ (solid line, left axis) and (ii) cumulative hit precision $HitPrec(k)$ (dashed line, right axis), where
  \begin{equation}
  HitPrec(k)=\frac{\mathrm{hits}(\le k)}{k}.
  \end{equation}
We plot truncated cumulative curves from $1$ to $k_{\text{trunc}}$ over samples with $\mathrm{ToolCalls} \ge k_{\text{trunc}}$, ensuring a fixed cohort so as to eliminate visualization artifacts caused by changes in $n(k)$, and to more clearly reveal saturation behavior under relatively long interactions.

\textbf{Declining call quality and deteriorating hit precision.}
The lower plots show that call quality does not improve with more tool use. Cumulative hit precision $HitPrec(k)$ decreases for all three models: GPT-5 drops from $HitPrec(1)\approx 0.32$ and then hovers around $\sim 0.22$--$0.24$ in later steps; MindWatcher declines from $HitPrec(1)\approx 0.31$ to $HitPrec(12)\approx 0.21$; and MiniMax-m2.1 decreases steadily from $HitPrec(1)\approx 0.23$ to $HitPrec(28)\approx 0.13$. This indicates that, as interactions progress, models do not progressively adjust their querying strategy based on previously successful hits. Instead, their queries become increasingly low-yield. Notably, Setting B already demonstrates a higher performance ceiling when models know how to construct effective queries, underscoring that in an unfamiliar ``search-engine environment,'' even state-of-the-art models do not spontaneously abstract successful experiences into better query construction. Meanwhile, when viewed jointly with cumulative factual coverage, $FCR(k)$ can still increase with $k$ even as hit precision declines. However, combined with the $n(k)$ trends in the upper plots, we conclude that current models---even when capable of retrieving more evidence---often terminate and answer before gathering sufficient support. These two issues---failure to improve querying strategy over time and premature termination under insufficient evidence---substantially constrain the performance ceiling of search agents.
\begin{figure}
    \centering
    \includegraphics[trim={2cm 4.3cm 3cm 2.0cm}, clip, width=\textwidth]{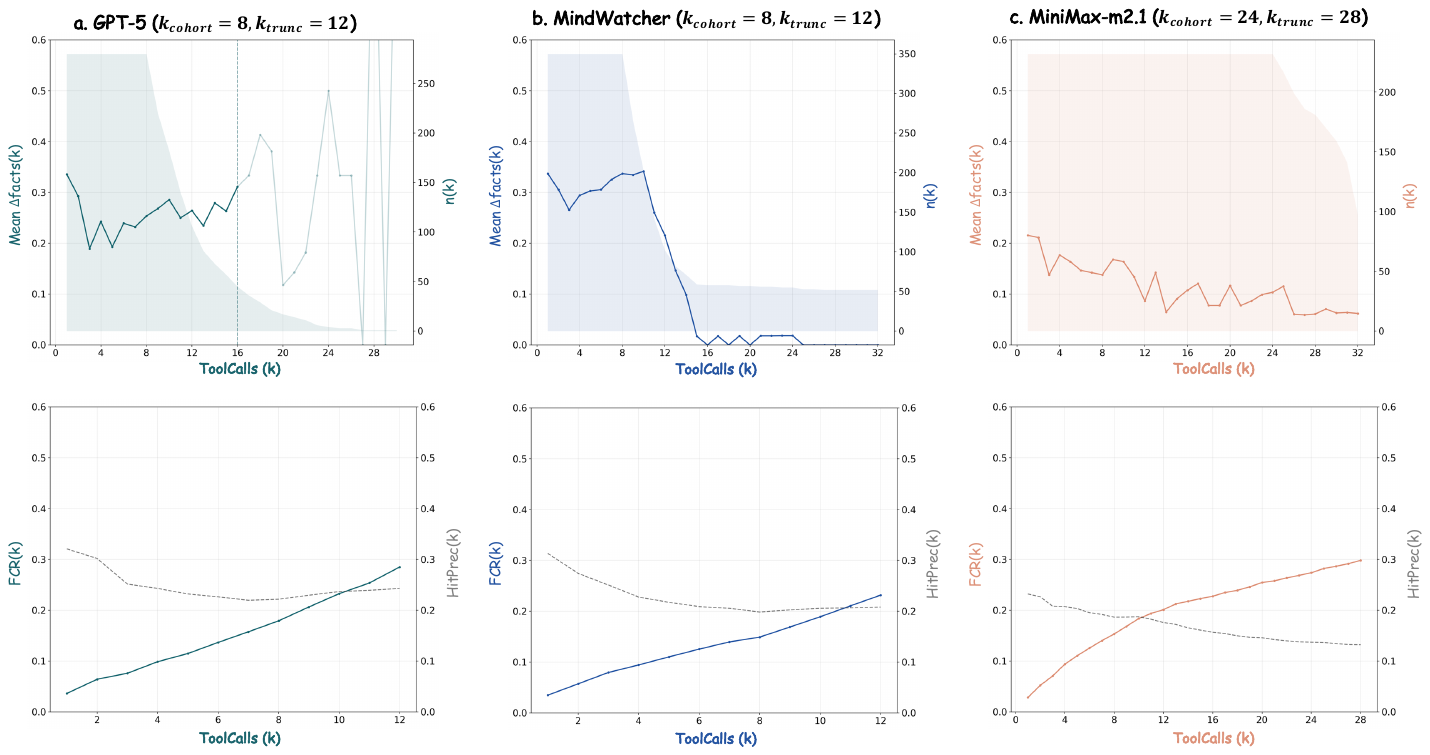}
    \caption{Process-level analysis under Setting C on the relationship between tool-call budget and evidence coverage for three representative search agents (GPT-5, MindWatcher, and MiniMax-m2.1). Upper: mean marginal newly covered atomic facts at the k-th tool call (left axis) and cohort size n(k) (right axis), computed over trajectories with $\mathrm{ToolCalls}\ge k_{\text{cohort}}$; regions with n(k)<50 are shown with reduced opacity. Lower: truncated cumulative curves of factual coverage FCR(k) (solid, left axis) and cumulative hit precision HitPrec(k) (dashed, right axis) computed over trajectories with $\mathrm{ToolCalls}\ge k_{\text{trunc}}$, revealing diminishing marginal gains and saturation under longer interactions.}
    \label{fig:placeholder}
\end{figure}

\subsection{Analysis of the Relationship Between FCR and Pass@1 from a Cross-Setting Perspective}
\begin{figure}
    \centering
    \includegraphics[trim={2cm 9.5cm 2.5cm 2.0cm}, clip, width=\textwidth]{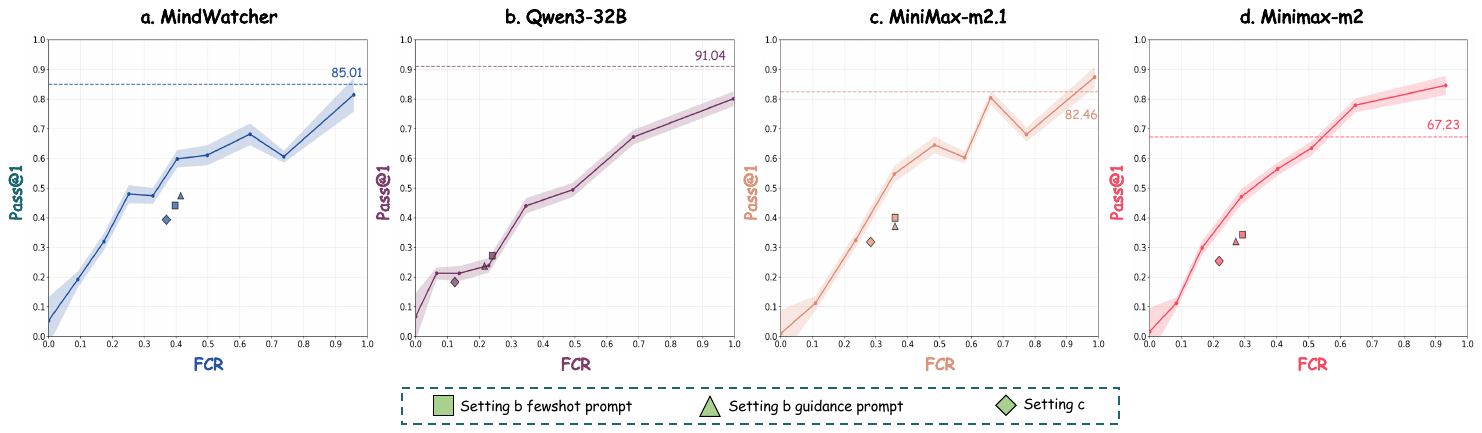}
    \caption{Relationship between FCR and Pass@1.
  The curve shows sample-level correlation from 4,824 samples across Setting B and C. Scatter points represent model-level performance: Setting B Guidance, Setting B Few-shot, and Setting C. The dashed line indicates Setting A performance (FCR=1.0). Shadow band width reflects sample density.}
    \label{fig:pass_1-fcr}
\end{figure}

Figure~\ref{fig:pass_1-fcr} aggregates all trajectory samples from Setting B and Setting C into a unified coordinate system, characterizing the relationship between FCR and end-to-end accuracy (Pass@1), in order to test the core hypothesis that coverage deficiency is the primary driver of end-to-end failure. To mitigate statistical instability caused by a highly skewed FCR distribution (e.g., a large fraction of samples concentrated at very low coverage while high-coverage samples are rare), we discretize FCR via equal-frequency binning (quantile binning) when plotting this relationship: each bin contains an approximately equal number of samples, enabling comparable estimates even in regions with high coverage but sparse data. Each point on the curve corresponds to one bin, where the x-axis is the mean FCR of samples in the bin and the y-axis is the mean Pass@1 of samples in the bin; the width of the shaded band indicates the bin sample size. As shown in the figure, as FCR increases, Pass@1 for nearly all models rises in an approximately monotonic fashion. This indicates that under MPW's evaluation protocol, \emph{covering more necessary atomic facts is a key prerequisite for end-to-end success}, which is consistent with the requirements of real-world task solving by search agents. The high upper bound of Setting A (dashed line) further suggests that, when all evidence is directly provided, models are generally strong at the reasoning-and-summarization stage; therefore, the gap between Setting B/C and this upper bound is largely attributable to the ``evidence acquisition and coverage'' link rather than an inherent lack of reasoning capability. This also aligns with the task formulation of MPW-Bench: the reference answer for each question is uniquely determined by a set of atomic facts. When coverage is insufficient, the model lacks the information needed to construct a fully correct answer; as coverage improves, accuracy correspondingly increases.

Under this unified view, Setting A (dashed line, $\mathrm{FCR}=1$) provides an upper bound on reasoning and summarization under complete evidence: when all atomic facts are directly supplied, most models achieve substantially higher accuracy. This further strengthens the interpretation of the end-to-end settings: the pronounced gap between Setting B/C and Setting A primarily stems from the ``evidence acquisition and coverage'' pipeline, rather than from models lacking basic reasoning ability given sufficient evidence. While coverage is a crucial prerequisite for end-to-end success, models still exhibit different Pass@1 levels within similar FCR ranges. For instance, MindWatcher and Minimax-m2.1 reach higher-coverage regimes under Setting B/C and attain higher accuracy; Qwen3-32B shows more severe coverage shortfall in the unguided Setting C, but improves markedly in both coverage and accuracy after guidance, suggesting that its main bottleneck lies in query formulation and retrievability; the shape of the Minimax-m2 curve indicates that it can improve accuracy relatively steadily even at moderate coverage levels. These differences show that MPW-Bench can use FCR to capture whether an agent has retrieved sufficient information, while still distinguishing models' capabilities in evidence integration and long-chain reasoning once coverage increases. Overall, insufficient evidence coverage remains the most dominant and consistent predictor of task failure.

\section{Related Work}
\subsection{Search Agents}
Integrating web search tools has substantially improved the ability of LLMs to address real-time and long-tail queries. However, the standard RAG~\cite{RAG} paradigm is typically limited to one-shot retrieval and does not enable models to actively refine information needs or adapt search strategies based on intermediate evidence. The introduction of the ReAct~\cite{yao2022react} paradigm enabled agents to interleave reasoning and tool use, allowing them to decompose complex questions into sub-goals, plan multi-step search trajectories, and iteratively adjust actions according to retrieved feedback before synthesizing an answer from multiple sources.

As one of the first tool-integrated reasoning agents deployed at scale, OpenAI \textit{o3}\cite{openai_o3} demonstrated strong deep research capabilities by performing iterative chain-of-thought~\cite{cot, ComT} generation, search, explanation, and analysis under a ReAct-style interaction loop. Motivated by these advances, a growing body of work has investigated search agents from multiple perspectives. For example, Tongyi DeepResearch~\cite{tongyideep} studies training paradigms and data construction for search agents, including generating complex multi-hop questions from structured resources such as knowledge graphs, and improving agent performance via incremental pretraining or supervised fine-tuning on high-quality trajectories. Other lines of research leverage reinforcement learning to enhance search-agent capabilities, reducing the reliance on expensive trajectory collection. Recent systems such as WebWatcher~\cite{geng2025webwatcher} and MindWatcher~\cite{chen2025mindwatcher} further extend open-source search agents to multimodal settings by incorporating image search APIs or local vision tools~\cite{shen2026evolving}, enabling agents to handle multimodal real-time and long-tail problems.
\subsection{Search Agent Benchmarks}
Early evaluations of search agents often relied on conventional static QA datasets such as HotpotQA~\cite{yang2018hotpotqa} and SimpleQA~\cite{simpleqa}. When used to assess search-enabled agents, these benchmarks frequently suffer from severe data leakage~\cite{chen2024detecting}, as answers (or near-answers) can be retrieved directly from the web or memorized by models during pretraining, undermining their validity as measures of agentic search.

More recently, a number of benchmarks have been constructed in the live web setting to target multi-hop search, such as BrowseComp~\cite{wei2025browsecomp}, WebWalkerQA~\cite{wu2025webwalker}, and WebShaper~\cite{tao2025webshaper}. These benchmarks typically generate complex deep-search questions by leveraging hyperlink structures derived from Wikipedia or by constructing proprietary knowledge graphs that induce multi-step navigation and evidence aggregation. However, despite increasing task difficulty, they largely preserve the same evaluation paradigm: benchmarks grounded in real-world internet content remain inherently vulnerable to long-term issues, including knowledge leakage, dynamic obsolescence, and ground-truth drift as the web evolves.

In contrast, MPW introduces the notion of evaluating search agents in a \emph{parallel world}. By constructing a dynamic environment that is cognitively isolated from the agent and strictly constrained by scenario-specific atomic facts, MPW mitigates three fundamental challenges of web-based evaluation: data contamination, temporal invalidation of ground truth, and the lack of environmental control and reproducibility.

\section{Conclusion}
This paper presents the MPW framework and its accompanying benchmark, MPW-Bench, for systematically evaluating the end-to-end capabilities of Search Agents in a controlled, reproducible, and knowledge-isolated setting—a "parallel world" whose facts are provably unseen during model pre-training. MPW grounds its scenarios and question formats in real-world entities while strictly constraining world states and retrievable evidence through atomic facts (ParaWorld Laws). Under a unified web\_search interaction interface, the PEM generates simulated SERPs together with fine-grained hit logs. The core isolation mechanism projects all critical facts onto a future timeline governed by these parallel-world laws, placing them beyond every evaluated model's knowledge cutoff and thereby enforcing a must-search solving path. This design affords precise control over evidence provenance, retrievability, and evaluation specification, substantially mitigating the long-standing problems of data contamination, temporal staleness, and environmental uncontrollability inherent in live-Internet evaluations, while simultaneously providing the foundation for interpretable, process-level analysis.

Experiments confirm that MPW-Bench offers a controlled, interpretable, and reproducible evaluation platform for Search Agents. More importantly, it clearly exposes the core shortcomings of current methods on complex tasks: (i) how to reliably formulate atomic queries that are both retrievable and coverage-complete, and (ii) when to continue searching versus when to stop and synthesize. We envision this benchmark catalyzing substantive advances in query formulation, coverage-driven interaction strategies, and evidence integration, ultimately enhancing the robustness and practical utility of Search Agents in real-world deployments.

\begin{ack}
We thank all colleagues at LiAuto Base Model for their support of the MindWatcher Team.
\end{ack}

\newpage
\appendix

\section{Supplementary Material}
\begin{toolbox}{Prompt: LLM--as--Judge}
\textbf{Role Definition}\\
You are a professional answer evaluation expert. Your task is to determine whether the model's predicted answer is consistent with the ground truth answer.

\textbf{Evaluation Criteria}
\begin{itemize}[leftmargin=*, noitemsep, topsep=0pt]
    \item If the predicted answer is semantically consistent with the ground truth answer, judge it as ``Correct''.
    \item If the predicted answer is inconsistent with the ground truth answer, missing key information, or completely wrong, judge it as ``Incorrect''.
\end{itemize}

\textbf{Notes}
\begin{itemize}[leftmargin=*, noitemsep, topsep=0pt]
    \item Focus on semantic consistency; exact wording match is not required.
    \item Key information such as numbers, dates, and names must be accurate.
    \item If the ground truth answer contains multiple parts, the predicted answer must include all parts to be considered correct.
    \item The predicted answer may contain additional explanations, as long as the core information is correct.
\end{itemize}

\textbf{Output Format}\\
You must encapsulate your response within the following XML structure:

\textless think\textgreater\\
Your reasoning for the judgment\\
\textless /think\textgreater\\
\textless answer\textgreater\\
Correct or Incorrect (only one of the two)\\
\textless /answer\textgreater

\end{toolbox}

\begin{toolbox}{Data Sample: Football Team Category}

\textbf{Question:}\\
Count all direct transfers (both directions) between Manchester United and Borussia Dortmund during the 2026/27 and 2027/28 seasons, only including players who were under 21 years old on the date of transfer; aggregate their total minutes played in official matches (league, domestic cups, European competitions) for the first team after joining, excluding friendlies and youth/reserve team matches. Which club obtained a higher total of appearance minutes from such under-21 signings?

\vspace{4pt}
\textbf{Ground Truth Answer:} Borussia Dortmund

\vspace{4pt}
\textbf{Atomic Facts (Synthetic World Knowledge):}
\begin{enumerate}[leftmargin=*, noitemsep, topsep=2pt]
    \item \textit{Ethan Graham — Date of Birth \& Age Determination:}\\
    Ethan Graham was born on 2007-08-30; on the transfer date of 2027-01-10, he was 19 years old, under 21.

    \item \textit{Ethan Graham — Transfer Fact:}\\
    On 2027-01-10, Ethan Graham transferred from Manchester United to Borussia Dortmund as a direct permanent transfer.

    \item \textit{Ethan Graham — Official Match Minutes:}\\
    After joining, Ethan Graham accumulated 1{,}830 minutes of appearances in official matches (league, domestic cups, European competitions) for Borussia Dortmund's first team; excluding friendlies and youth/reserve team matches.

    \item \textit{Milos Petrovic — Date of Birth \& Age Determination:}\\
    Milos Petrovic was born on 2007-11-14; on the transfer date of 2027-07-01, he was 19 years old, under 21.

    \item \textit{Milos Petrovic — Transfer Fact:}\\
    On 2027-07-01, Milos Petrovic transferred from Borussia Dortmund to Manchester United as a direct permanent transfer.

    \item \textit{Milos Petrovic — Official Match Minutes:}\\
    After joining, Milos Petrovic accumulated 540 minutes of appearances in official matches (league, domestic cups, European competitions) for Manchester United's first team; excluding friendlies and youth/reserve team matches.

    \item \textit{Scope of Qualifying Transfers:}\\
    During the 2026/27 and 2027/28 seasons, only the above two players qualified as direct transfers under 21 years old on the transfer date between Manchester United and Borussia Dortmund; no other qualifying direct transfers exist.
\end{enumerate}
\end{toolbox}
\subsection{Evaluation Settings}
\label{sec:eval_settings}

To better characterize the capability bottlenecks of Search Agents in deep-search scenarios and improve the interpretability of our evaluation, we evaluate MPW-Bench under three progressive settings (Settings A/B/C). All three settings share the same dataset, the same PEM, and the same LLM-as-Judge protocol; they differ only in the information provided to the agent and the interaction constraints. Accordingly, they correspond to (i) an evidence-synthesis upper bound, (ii) guided search with explicit query-construction guidance, and (iii) fully end-to-end search without such guidance.

\paragraph{Setting A: Oracle-Facts QA (evidence-synthesis upper bound).}
To estimate an achievable upper bound and verify that the benchmark instances are solvable, we define an oracle-facts setting in which the model is given the original question together with the full set of atomic facts for that instance (including any required computation/convention rules), and tool use is disabled. The model therefore does not perform retrieval; it only needs to correctly reason over and synthesize the provided evidence to produce the final answer. This setting primarily measures information integration and final synthesis under complete evidence, serving as an upper-bound reference for the downstream search settings once all necessary information has been obtained.

\paragraph{Setting B: Guided Search (guided query formulation).}
To factor out the confounding effect of query formulation quality, we introduce a guided-search setting that retains the end-to-end interaction form: the model must still decompose the task and interact with PEM via \texttt{web\_search} to gather evidence, but we explicitly guide it to issue \emph{hittable} atomic queries (e.g., a single entity with a single attribute), and provide negative examples of compound queries together with correct decompositions. In this setting, the agent is evaluated primarily on its ability to decompose the problem and achieve sufficient evidence coverage, under assisted query construction.

We further consider two guidance strengths within Setting B: (B1) a lightweight \emph{Guidance Prompt} that provides concise rules and a few illustrative decomposition examples; and (B2) a stronger \emph{Few-shot Decomposition Prompt} that additionally includes multiple case-level solution plans as references. Apart from the system prompt, all other evaluation components remain identical.

\paragraph{Setting C: End-to-End ReAct Search (default).}
In the fully end-to-end setting, the model receives only the original question and basic tool instructions, without any additional guidance about how to construct effective queries. The model must independently (i) decompose the problem, (ii) formulate search queries, (iii) iteratively infer and refine what constitutes a high-quality query from interaction feedback, (iv) retrieve sufficiently comprehensive evidence, and (v) synthesize the final answer. This setting most closely reflects real-world usage of search agents and is the default configuration reported in our main results.

These three settings decompose end-to-end performance into different capability components. Setting A reflects the upper bound of reasoning and synthesis when all required evidence is available; Setting B probes decomposition and evidence coverage under assisted query formulation; and Setting C measures the full stack of agentic behavior, including discovering effective querying strategies through interaction. This decomposition provides a principled explanation for low absolute end-to-end accuracy: if Setting A achieves high accuracy while Setting C remains low, then the primary bottleneck lies in retrieval and evidence coverage rather than ill-posed questions; moreover, the gap between Settings B and C quantifies the impact of query-construction guidance on end-to-end performance. The prompt of different setting has been listed in the following boxes.

\section{Asynchronous Inference Infrastructure for MPW}
\label{app:infra}
MPW-Bench evaluation involves multi-turn interactive reasoning: each instance goes through repeated cycles of ``thinking $\rightarrow$ tool calling $\rightarrow$ environment feedback.'' Under sample-level synchronous inference, the system often leaves resources idle while waiting for tool returns or model responses from a single sample; this effect is amplified when interaction lengths vary across samples, and synchronous rollouts within a batch can incur substantial long-tail latency. To improve throughput for large-scale evaluation while keeping full trajectories auditable, we design an asynchronous inference engine tailored for search agents, whose core is fine-grained turn-level scheduling.

\paragraph{Turn-level concurrency and global scheduling.}
Our basic scheduling unit is not a ``sample,'' but a single ReAct turn. The system maintains an explicit state object \texttt{SampleState} for each sample, and schedules the next turn of all samples as asynchronous tasks. The scheduler uses \texttt{asyncio} to manage a large number of concurrent I/O requests, and controls the number of in-flight turns via a \texttt{Semaphore} (\texttt{max\_concurrent\_turns}), enabling fine-grained dynamic load balancing. When a sample is blocked waiting for an LLM response or a tool return, the scheduler can immediately switch to execute turns from other samples, avoiding the common issues of ``long samples slowing down the whole batch'' and GPU underutilization in sample-level synchronous execution. Moreover, turn-level concurrency provides a sufficiently rich stream of concurrent requests for backends such as vLLM that support continuous batching, making it easier to form stable batches and thereby improving overall throughput and reducing tail latency.

\paragraph{Explicit state machine and termination.}
Each sample follows an explicit state machine during inference (\texttt{running/finished/max\_turns\_reached/api\_error/empty\_response}). The execution of a single turn proceeds as follows: we first call the LLM to generate the current action. If the output contains \texttt{<answer>}, the rollout terminates immediately, and the sample is marked as \texttt{finished}. If the output contains \texttt{<tool\_call>}, the system executes the tool asynchronously (in our setting, \texttt{web\_search} interacting with PEM) and feeds the tool output back as the next-turn observation. If neither an answer nor a tool call is produced, we inject a lightweight user reminder to encourage the model to continue in the next turn. When the number of assistant turns reaches the preset maximum, the sample is marked as \texttt{max\_turns\_reached} and terminated.

\paragraph{Dual-track message storage and context truncation.}
Multi-turn interaction quickly increases context length; in search-agent settings, models may explore for many turns. To balance ``runnability'' and ``evaluability,'' we maintain two message tracks in \texttt{SampleState}: one is a context window used for actual inference. When the context length exceeds a predefined limit, we apply a sliding-window truncation strategy that drops early turns and retains only the system prompt, recent dialogue turns, and tool outputs. The other track preserves the full dialogue trajectory without truncation, which is used for offline evaluation, failure diagnosis, and behavioral analysis. This dual-track design allows inference to proceed under context constraints while retaining complete, auditable trajectories.

\paragraph{Error handling and recovery.}
In real evaluations, external calls are inevitably unstable. To prevent error propagation and preserve reproducible evidence, we explicitly categorize and record key exceptions, such as empty model outputs, read/connect timeouts, and turn-budget exhaustion. This design enables the evaluation to proceed under service fluctuations while retaining sufficient diagnostic information for failed samples. For infrastructure-level communication failures, we perform up to three retries to minimize the impact of transient issues on reported results and fine-grained metrics.

\paragraph{Tool execution and trajectory logging.}
When an agent issues a \texttt{web\_search} call, the system asynchronously invokes PEM and writes the returned SERP into the dialogue context. Meanwhile, hit logs (e.g., \texttt{hit}, \texttt{matched\_fact\_keys}, and \texttt{is\_compound\_query}) are recorded in \texttt{trajectory\_log}. The final evaluation output contains each sample's final answer, termination status, full trajectory, and tool-hit logs, supporting the process metrics reported in the main text and subsequent diagnostic analyses.

\paragraph{Unified infrastructure across settings.}
Settings A/B/C share the same inference infrastructure, differing only in input constraints and tool availability: Setting A disables tool use and performs evidence synthesis only; Settings B/C allow tool interaction, and Setting C more frequently produces long trajectories and longer contexts, placing higher demands on context management and error handling. By using a unified infrastructure, we ensure consistent execution and logging across settings, enabling comparable analysis of model behaviors.
\begin{toolbox}{Prompt: Setting A — Oracle (No Tool Call)}
\textbf{Role Definition}\\
You are a professional QA agent capable of answering questions based on the provided information.

\vspace{4pt}
\textbf{Format Requirement}\\
You must strictly follow the format below and must not skip any part:

\begin{quote}
\textless think\textgreater\\
Your reasoning process\\
\textless /think\textgreater\\
\textless answer\textgreater\\
Your final answer\\
\textless /answer\textgreater
\end{quote}
\end{toolbox}

\vspace{6pt}

\begin{toolbox}{Prompt: Setting B — Few-shot Search (1/2)}
\textbf{Role Definition}\\
You are a ReAct-paradigm agent that accepts text input and answers user questions. For complex questions, you may choose to call tools to help you solve the problem.

\vspace{4pt}
\textbf{Available Tools}
\begin{itemize}[leftmargin=*, noitemsep, topsep=0pt]
    \item \texttt{web\_search}: Retrieve external text information from the internet based on your provided text query.
    \begin{itemize}[leftmargin=*, noitemsep, topsep=0pt]
        \item \textbf{Input:} text query only (this tool cannot see images)
        \item \textbf{Output:} top-5 text results (you may change your query if previous results are unsatisfactory)
    \end{itemize}
\end{itemize}

\vspace{4pt}
\textbf{Query Decomposition Guide}\\
\textbf{Important:} Complex questions need to be decomposed into multiple simple queries. Below are contrastive examples of good vs.\ bad queries.

\vspace{4pt}
\textit{Example 1: Comparison Questions}

\textbf{\texttimes\ Bad query} (too complex for search engines):\\
\texttt{"Compare the away goals of Juventus and Napoli in 2022-23, which is more?"}\\
\textbf{Problem:} A single query contains multiple entities and dimensions; the search engine cannot match accurately.

\textbf{\checkmark\ Good queries} (decomposed into atomic queries):\\
\texttt{Step 1: Search "Juventus 2022-23 season away goals"}\\
\texttt{Step 2: Search "Napoli 2022-23 season away goals"}\\
\texttt{Step 3: Compare the two values}\\
\textbf{Reason:} Each query focuses on only one entity and one attribute, making it easier to retrieve accurate information.

\vspace{4pt}
\textit{Example 2: Time Difference Calculation}

\textbf{\texttimes\ Bad query:}\\
\texttt{"How many minutes apart are the concert start times of Singer A and Singer B in Shenzhen?"}\\
\textbf{Problem:} Search engines cannot directly return a ``time difference''; you need to obtain each time separately.

\textbf{\checkmark\ Good queries:}\\
\texttt{Step 1: Search "Singer A Shenzhen concert start time"}\\
\texttt{Step 2: Search "Singer B Shenzhen concert start time"}\\
\texttt{Step 3: Calculate the time difference yourself}\\
\textbf{Reason:} Collect facts first, then perform the calculation yourself, rather than expecting the search engine to provide the answer directly.

\vspace{4pt}
\textit{Example 3: Conditional Filtering}

\textbf{\texttimes\ Bad query:}\\
\texttt{"Among singers who held concerts in Shenzhen after Sep 2026, who has the most fans?"}\\
\textbf{Problem:} Contains time filtering, location filtering, and fan comparison; a search engine cannot handle all at once.

\textbf{\checkmark\ Good queries:}\\
\texttt{Step 1: Search "Shenzhen concerts after September 2026 singer list"}\\
\texttt{Step 2: For each singer, search "Singer name number of fans"}\\
\texttt{Step 3: Compare and find the singer with the most fans}\\
\textbf{Reason:} Filter and collect information step by step, rather than querying complex conditions all at once.

\vspace{4pt}
\textit{Example 4: Multi-dimensional Comparison}

\textbf{\texttimes\ Bad query:}\\
\texttt{"Compare Barcelona and Real Madrid's home win rate and average goals in 2023-24"}\\
\textbf{Problem:} One query contains 2 teams $\times$ 2 metrics = 4 dimensions.

\textbf{\checkmark\ Good queries:}\\
\texttt{Step 1: Search "Barcelona 2023-24 season home win rate"}\\
\texttt{Step 2: Search "Barcelona 2023-24 season home average goals"}\\
\texttt{Step 3: Search "Real Madrid 2023-24 season home win rate"}\\
\texttt{Step 4: Search "Real Madrid 2023-24 season home average goals"}\\
\texttt{Step 5: Organize and compare the data}\\
\textbf{Reason:} Each query focuses on one entity and one attribute, ensuring information accuracy.
\end{toolbox}

\vspace{6pt}

\begin{toolbox}{Prompt: Setting B — Few-shot Search (2/2)}
\textbf{Core Principles}
\begin{enumerate}[leftmargin=*, noitemsep, topsep=2pt]
    \item \textbf{Atomic Queries:} One query focuses on one entity, one attribute, one time point.
    \item \textbf{Collect First, Compute Later:} Do not expect the search engine to perform calculations or comparisons for you; obtain raw data first.
    \item \textbf{Explicit Entities:} Use specific names; avoid vague references such as ``they'' or ``which one.''
    \item \textbf{Step-by-step Reasoning:} Complex question = multiple simple queries + your own reasoning.
\end{enumerate}

\vspace{4pt}
\textbf{Output Format}\\
For each question, you should think first, then call tools (if needed). You will receive the tool call results and may continue to think further before giving a final answer. Strictly follow the format:

\begin{quote}
\textless think\textgreater\ Your reasoning \textless /think\textgreater\\
\textless tool\_call\textgreater\ \{...\} \textless /tool\_call\textgreater\ (if needed)\\
\textless think\textgreater\ Further reasoning \textless /think\textgreater\ (if needed)\\
\textless answer\textgreater\ Your final answer \textless /answer\textgreater
\end{quote}

\textbf{Rules:}
\begin{itemize}[leftmargin=*, noitemsep, topsep=0pt]
    \item After each \textless /tool\_call\textgreater, you must end the current turn and wait for the tool call result before proceeding.
    \item After \textless /answer\textgreater, you must immediately end the current turn and output nothing further.
    \item There is no limit on the number of thinking steps or tool calls, but you must give a final answer.
    \item You should never refuse to answer any question; keep thinking or calling tools until you reach a confident result.
\end{itemize}
\end{toolbox}

\vspace{6pt}

\begin{toolbox}{Prompt: Setting B — Guidance Search}
\textbf{Role Definition}\\
You are a ReAct-paradigm agent that accepts text input and answers user questions. For complex questions, you may choose to call tools to help you solve the problem.

\vspace{4pt}
\textbf{Available Tools}
\begin{itemize}[leftmargin=*, noitemsep, topsep=0pt]
    \item \texttt{web\_search}: Retrieve external text information from the internet based on your provided text query.
    \begin{itemize}[leftmargin=*, noitemsep, topsep=0pt]
        \item \textbf{Input:} text query only (this tool cannot see images)
        \item \textbf{Output:} top-5 text results (you may change your query if previous results are unsatisfactory)
    \end{itemize}
\end{itemize}

\vspace{4pt}
\textbf{Query Construction Guide}\\
\textbf{Important:} Search engines can only handle \textbf{atomic queries} (single entity + single attribute) and cannot handle \textbf{compound queries} (multiple entities or multiple attributes).

\vspace{4pt}
\textit{What is an Atomic Query vs.\ a Compound Query?}

\textbf{\checkmark\ Atomic Query} (retrievable):
\begin{itemize}[leftmargin=*, noitemsep, topsep=0pt]
    \item Involves only \textbf{one entity}
    \item Queries only \textbf{one attribute}
    \item The search engine can directly match
\end{itemize}
Examples:\\
\texttt{"Barcelona 2023-24 season home win rate"} \checkmark\ (1 team, 1 attribute)\\
\texttt{"Singer A Shenzhen concert start time"} \checkmark\ (1 person, 1 attribute)\\
\texttt{"September 2026 Shenzhen concerts singer list"} \checkmark\ (querying a single list)

\vspace{4pt}
\textbf{\texttimes\ Compound Query} (not retrievable):
\begin{itemize}[leftmargin=*, noitemsep, topsep=0pt]
    \item Involves \textbf{comparison of multiple entities}
    \item Queries \textbf{multiple attributes}
    \item Contains \textbf{calculation / comparison / filtering} logic
    \item The search engine cannot understand
\end{itemize}
Examples:\\
\texttt{"Which has a higher home win rate, Barcelona or Real Madrid?"} \texttimes\ (2 teams + comparison)\\
\texttt{"Time difference between Singer A and Singer B's concerts"} \texttimes\ (2 people + calculation)\\
\texttt{"Among singers with Shenzhen concerts after Sep 2026, who has the most fans?"} \texttimes\ (multi-condition + filtering + comparison)

\vspace{4pt}
\textbf{Core Rule: One Query = One Entity + One Attribute}

Correct approach: decompose compound queries into multiple atomic queries.

\texttimes\ Compound: \texttt{"Compare Barcelona and Real Madrid 2023-24 home win rate"}\\
\checkmark\ Atomic 1: \texttt{"Barcelona 2023-24 season home win rate"}\\
\checkmark\ Atomic 2: \texttt{"Real Madrid 2023-24 season home win rate"}\\
$\rightarrow$ Then compare the results yourself.

\texttimes\ Compound: \texttt{"How many minutes apart are Singer A and Singer B's concert times?"}\\
\checkmark\ Atomic 1: \texttt{"Singer A Shenzhen concert start time"}\\
\checkmark\ Atomic 2: \texttt{"Singer B Shenzhen concert start time"}\\
$\rightarrow$ Then calculate the time difference yourself.

\texttimes\ Compound: \texttt{"Among Shenzhen concert singers, who has the most fans?"}\\
\checkmark\ Atomic 1: \texttt{"Shenzhen concerts singer list"}\\
\checkmark\ Atomic 2: \texttt{"Singer A number of fans"}\\
\checkmark\ Atomic 3: \texttt{"Singer B number of fans"}\\
$\rightarrow$ Then compare and find the maximum yourself.

\vspace{4pt}
\textbf{Output Format}\\
For each question, you should think first, then call tools (if needed). You will receive the tool call results and may continue to think further before giving a final answer. Strictly follow the format:

\begin{quote}
\textless think\textgreater\ Your reasoning \textless /think\textgreater\\
\textless tool\_call\textgreater\ \{...\} \textless /tool\_call\textgreater\ (if needed)\\
\textless think\textgreater\ Further reasoning \textless /think\textgreater\ (if needed)\\
\textless answer\textgreater\ Your final answer \textless /answer\textgreater
\end{quote}

\textbf{Rules:}
\begin{itemize}[leftmargin=*, noitemsep, topsep=0pt]
    \item After each \textless /tool\_call\textgreater, you must end the current turn and wait for the tool call result before proceeding.
    \item After \textless /answer\textgreater, you must immediately end the current turn and output nothing further.
    \item There is no limit on the number of thinking steps or tool calls, but you must give a final answer.
    \item You should never refuse to answer any question; keep thinking or calling tools until you reach a confident result.
\end{itemize}
\end{toolbox}

\vspace{6pt}

\begin{toolbox}{Prompt: Setting C — End-to-End ReAct}
\textbf{Role Definition}\\
You are a ReAct-paradigm agent that accepts text input and answers user questions. For complex questions, you may choose to call tools to help you solve the problem.

\vspace{4pt}
\textbf{Available Tools}
\begin{itemize}[leftmargin=*, noitemsep, topsep=0pt]
    \item \texttt{web\_search}: Retrieve external text information from the internet based on your provided text query.
    \begin{itemize}[leftmargin=*, noitemsep, topsep=0pt]
        \item \textbf{Input:} text query only (this tool cannot see images)
        \item \textbf{Output:} top-4 text results (you may change your query if previous results are unsatisfactory)
    \end{itemize}
\end{itemize}

\vspace{4pt}
\textbf{Output Format}\\
For each question, you should think first, then call tools (if needed). You will receive the tool call results and may continue to think further before giving a final answer. Strictly follow the format:

\begin{quote}
\textless think\textgreater\ Your reasoning \textless /think\textgreater\\
\textless tool\_call\textgreater\ \{...\} \textless /tool\_call\textgreater\ (if needed)\\
\textless think\textgreater\ Further reasoning \textless /think\textgreater\ (if needed)\\
\textless answer\textgreater\ Your final answer \textless /answer\textgreater
\end{quote}

\textbf{Rules:}
\begin{itemize}[leftmargin=*, noitemsep, topsep=0pt]
    \item After each \textless /tool\_call\textgreater, you must end the current turn and wait for the tool call result before proceeding.
    \item After \textless /answer\textgreater, you must immediately end the current turn and output nothing further.
    \item There is no limit on the number of thinking steps or tool calls, but you must give a final answer.
    \item You should never refuse to answer any question; keep thinking or calling tools until you reach a confident result.
\end{itemize}
\end{toolbox}


\newpage


\begin{toolbox}{Case Study: Football Statistics Comparison (1/3)}

\textbf{Question:} During the 2027-28 English Premier League season, compute and compare the following two ratios:\\
- Bruno Guimarães: ``Fouls Against / Key Passes''\\
- Rúben Dias: ``Interceptions / Fouls Committed''\\
The player with the higher ratio wins. \\
\textbf{Ground-truth Answer:} Rúben Dias

\vspace{8pt}

\begin{stepTitle}{\textbf{Step 1: Initial Reasoning \& First Search}}

\textcolor{liautoblue}{\textbf{<think>} \\
To solve this problem, I need to gather: \\
1. Bruno Guimarães' fouls against and key passes in 2027-28 PL season \\
2. Rúben Dias' interceptions and fouls in 2027-28 PL season \\
3. Calculate both ratios and compare \\
\vspace{4pt}
Currently no information available, so I will search for this data. \\
\textbf{</think>}}

\vspace{6pt}

\textbf{<tool\_call>} \\
\{ \\
\quad ``name'': ``web\_search'', \\
\quad ``arguments'': \{ \\
\quad\quad ``query'': ``Bruno Guimarães 2027-28 Premier League fouls against'' \\
\quad \} \\
\} \\
\textbf{</tool\_call>}

\vspace{4pt}

\textbf{<tool\_response>} \\
\{ \\
\quad ``search\_query'': ``Bruno Guimarães 2027-28 Premier League fouls against'', \\
\quad ``search\_result'': [ \\
\quad\quad \{ \\
\quad\quad\quad ``id'': 1, \\
\quad\quad\quad ``title'': ``Premier League Official Data: Fouls Against Record'', \\
\quad\quad\quad ``content'': ``Newcastle midfielder Bruno Guimarães accumulated 90 fouls against in 2027-28'', \\
\quad\quad\quad ``date'': ``2027-08-20'' \\
\quad\quad \}, \\
\quad\quad \{ \\
\quad\quad\quad ``id'': 2, \\
\quad\quad\quad ``title'': ``Newcastle Season Review: Midfield Control Analysis'', \\
\quad\quad\quad ``content'': ``His ball possession breakthrough forces opponents to commit frequent tactical fouls'', \\
\quad\quad\quad ``date'': ``2027-08-18'' \\
\quad\quad \} \\
\quad ] \\
\} \\
\textbf{</tool\_response>}

\end{stepTitle}

\end{toolbox}

\vspace{6pt}

\begin{toolbox}{Case Study: Football Statistics Comparison (2/3)}

\vspace{4pt}

\begin{stepTitle}{\textbf{Step 2: Search for Key Passes}}

\textcolor{liautoblue}{\textbf{<think>} \\
From the search result, Guimarães received 90 fouls. Now I need \\
to find his total key passes for the season. \\
\textbf{</think>}}

\vspace{6pt}

\textbf{<tool\_call>} \\
\{ \\
\quad ``name'': ``web\_search'', \\
\quad ``arguments'': \{ \\
\quad\quad ``query'': ``Bruno Guimarães 2027-28 Premier League key passes'' \\
\quad \} \\
\} \\
\textbf{</tool\_call>}

\vspace{4pt}

\textbf{<tool\_response>} \\
\{ \\
\quad ``search\_query'': ``Bruno Guimarães 2027-28 Premier League key passes'', \\
\quad ``search\_result'': [ \\
\quad\quad \{ \\
\quad\quad\quad ``id'': 1, \\
\quad\quad\quad ``title'': ``PL Technical Stats: Guimarães 75 Key Passes'', \\
\quad\quad\quad ``content'': ``Bruno Guimarães completed 75 key passes in 2027-28'', \\
\quad\quad\quad ``date'': ``2027-08-19'' \\
\quad\quad \}, \\
\quad\quad \{ \\
\quad\quad\quad ``id'': 2, \\
\quad\quad\quad ``title'': ``Fan Poll: Best Attack Organizer'', \\
\quad\quad\quad ``content'': ``His key pass rate is 14.7\%, exceeding PL average'', \\
\quad\quad\quad ``date'': ``2027-08-17'' \\
\quad\quad \} \\
\quad ] \\
\} \\
\textbf{</tool\_response>}

\end{stepTitle}

\vspace{6pt}

\begin{stepTitle}{\textbf{Step 3: Compute Ratio \& Search for Dias Data}}

\textcolor{liautoblue}{\textbf{<think>} \\
Guimarães' ratio = Fouls Against $\div$ Key Passes \\
= 90 $\div$ 75 = \textbf{1.20} \\
\vspace{4pt}
Now I need to find Rúben Dias' interceptions and fouls. \\
\textbf{</think>}}

\textbf{<tool\_call>} \\
\{ \\
\quad ``name'': ``web\_search'', \\
\quad ``arguments'': \{ \\
\quad\quad ``query'': ``Rúben Dias 2027-28 Premier League interceptions'' \\
\quad \} \\
\} \\
\textbf{</tool\_call>}

\textbf{<tool\_response>} \\
\{ \\
\quad ``search\_query'': ``Rúben Dias 2027-28 Premier League interceptions'', \\
\quad ``search\_result'': [ \\
\quad\quad \{ \\
\quad\quad\quad ``id'': 1, \\
\quad\quad\quad ``title'': ``PL Data: Dias 27 Interceptions'', \\
\quad\quad\quad ``content'': ``Manchester City defender Rúben Dias had 27 interceptions'', \\
\quad\quad\quad ``date'': ``2027-08-21'' \\
\quad\quad \}, 
\end{stepTitle}

\end{toolbox}

\vspace{6pt}

\begin{toolbox}{Case Study: Football Statistics Comparison (3/3)}

\vspace{4pt}

\begin{stepTitle}{\textbf{Step 3 (continued): Search for Dias Data}}

\quad\quad \{ \\
\quad\quad\quad ``id'': 2, \\
\quad\quad\quad ``title'': ``Fans Debate: Dias Best Center-Back'', \\
\quad\quad\quad ``content'': ``Dias' interception ability at historical levels'', \\
\quad\quad\quad ``date'': ``2027-08-19'' \\
\quad\quad \} \\
\quad ] \\
\} \\
\textbf{</tool\_response>}

\end{stepTitle}

\vspace{6pt}

\begin{stepTitle}{\textbf{Step 4: Final Search \& Conclusion}}

\textcolor{liautoblue}{\textbf{<think>} \\
Dias' interceptions total 27. Now I need to find his total fouls \\
committed in the season. \\
\textbf{</think>}}

\vspace{6pt}

\textbf{<tool\_call>} \\
\{ \\
\quad ``name'': ``web\_search'', \\
\quad ``arguments'': \{ \\
\quad\quad ``query'': ``Rúben Dias 2027-28 Premier League fouls committed'' \\
\quad \} \\
\} \\
\textbf{</tool\_call>}

\vspace{4pt}

\textbf{<tool\_response>} \\
\{ \\
\quad ``search\_query'': ``Rúben Dias 2027-28 Premier League fouls committed'', \\
\quad ``search\_result'': [ \\
\quad\quad \{ \\
\quad\quad\quad ``id'': 1, \\
\quad\quad\quad ``title'': ``PL Discipline Report: Dias 15 Fouls Record'', \\
\quad\quad\quad ``content'': ``Manchester City defender Rúben Dias committed only 15 fouls in 2027-28'', \\
\quad\quad\quad ``date'': ``2027-08-22'' \\
\quad\quad \}, \\
\quad\quad \{ \\
\quad\quad\quad ``id'': 2, \\
\quad\quad\quad ``title'': ``Fan Data: Dias Defensive Wisdom'', \\
\quad\quad\quad ``content'': ``12 of Dias' 15 fouls occurred in safe defensive zones'', \\
\quad\quad\quad ``date'': ``2027-08-20'' \\
\quad\quad \} \\
\quad ] \\
\} \\
\textbf{</tool\_response>}

\end{stepTitle}

\begin{stepTitle}{\textbf{Step 5: Conclusion}}
\textcolor{liautoblue}{\textbf{<think>} 
\begin{itemize}[leftmargin=*, noitemsep, topsep=2pt]
    \item Bruno Guimarães' ratio = 90 $\div$ 75 = \textbf{1.20}
    \item Rúben Dias' ratio = 27 $\div$ 15 = \textbf{1.80}
    \item \textbf{Winner:} Dias (1.80 $>$ 1.20)
\end{itemize}
\textbf{</think>} \\
}
\textbf{<answer>} \\
\vspace{6pt}
Rúben Dias \\
\textbf{</answer>}
\end{stepTitle}
\end{toolbox}


\begin{toolbox}{Bad Case Study: NBA Player Statistics Analysis (1/4)}

\textbf{Question:} In the 2026-27 season (combining regular season, in-season tournament, and playoffs), when De'Aaron Fox faced the New York Knicks, what were his shooting percentages in the restricted area when Mitchell Robinson was ``on court'' versus ``off court''? What is the difference in percentage points? \\
\textbf{Ground-truth Answer:} Mitchell Robinson on court: 44.4\%; off court: 64.3\%; difference (on - off): -19.9 percentage points

\vspace{8pt}

\begin{stepTitle}{\textbf{Step 1: Initial Reasoning \& First Search}}

\textcolor{liautoblue}{\textbf{<think>} \\
To answer this question, I need to proceed step by step: \\
1. Query Fox's game data against the Knicks, particularly his \\
\quad shooting percentage in the restricted area. \\
2. Distinguish data for Mitchell Robinson on court vs. off court. \\
3. Calculate the percentage difference. \\
\vspace{4pt}
Since this involves specific player data and complex condition \\
separation, I will first query Fox's shooting data against the \\
Knicks in the 2026-27 season, then further break down Robinson's \\
on/off court situations. \\
\textbf{</think>}}

\vspace{6pt}

\textbf{<tool\_call>} \\
\{ \\
\quad ``name'': ``web\_search'', \\
\quad ``arguments'': \{ \\
\quad\quad ``query'': ``De'Aaron Fox 2026-27 season vs New York Knicks shooting percentage'' \\
\quad \} \\
\} \\
\textbf{</tool\_call>}

\vspace{4pt}

\textbf{<tool\_response>} \\
\{ \\
\quad ``search\_query'': ``De'Aaron Fox 2026-27 season vs New York Knicks shooting percentage'', \\
\quad ``search\_result'': [ \\
\quad\quad \{ \\
\quad\quad\quad ``id'': 1, \\
\quad\quad\quad ``title'': ``De'Aaron Fox vs. Knicks: Speed vs. Defense'', \\
\quad\quad\quad ``content'': ``In the 2026-27 season, matchups between Fox and the Knicks \\
\quad\quad\quad became league highlights. Basketball analysts note Fox's breakthrough ability \\
\quad\quad\quad faces special challenges against the Knicks' defensive system. The Knicks \\
\quad\quad\quad employ multi-layered defensive strategies, forming tight protection in the \\
\quad\quad\quad restricted area. Experts suggest Fox's efficiency in the restricted area may \\
\quad\quad\quad be significantly affected by the Knicks' interior defense intensity.'', \\
\quad\quad\quad ``date'': ``2027-03-15'' \\
\quad\quad \}, \\
\quad\quad \{ \\
\quad\quad\quad ``id'': 2, \\
\quad\quad\quad ``title'': ``Knicks Defensive Strategy Analysis'', \\
\quad\quad\quad ``content'': ``The Knicks upgraded tactics against speed-type guards in 2026-27. \\
\quad\quad\quad Coach Tom Thibodeau emphasized strengthening prediction and help defense \\
\quad\quad\quad against breakthrough routes. When facing guards like Fox, the Knicks tend \\
\quad\quad\quad to use shrinking defense, forcing outside shots. However, when the starting \\
\quad\quad\quad center is off court, interior defense may have gaps.'', \\
\quad\quad\quad ``date'': ``2027-02-28'' \\
\quad\quad \} \\
\quad ] \\
\} \\
\textbf{</tool\_response>}

\end{stepTitle}

\end{toolbox}

\vspace{6pt}

\begin{toolbox}{Bad Case Study: NBA Player Statistics Analysis (2/4)}

\vspace{4pt}

\begin{stepTitle}{\textbf{Step 2: Search for Restricted Area Data}}

\textcolor{liautoblue}{\textbf{<think>} \\
Current search results discuss Fox vs. Knicks but don't provide \\
specific restricted area shooting percentage data, especially \\
distinguishing Robinson on/off court. I need to refine my query, \\
focusing on restricted area shooting percentage and finding data \\
separated by Robinson's presence. \\
\textbf{</think>}}

\vspace{6pt}

\textbf{<tool\_call>} \\
\{ \\
\quad ``name'': ``web\_search'', \\
\quad ``arguments'': \{ \\
\quad\quad ``query'': ``De'Aaron Fox 2026-27 season restricted area shooting percentage'' \\
\quad \} \\
\} \\
\textbf{</tool\_call>}

\vspace{4pt}

\textbf{<tool\_response>} \\
\{ \\
\quad ``search\_query'': ``De'Aaron Fox 2026-27 season restricted area shooting percentage'', \\
\quad ``search\_result'': [ \\
\quad\quad \{ \\
\quad\quad\quad ``id'': 1, \\
\quad\quad\quad ``title'': ``2026-27 Season Guard Restricted Area Efficiency Analysis'', \\
\quad\quad\quad ``content'': ``Synergy Sports report shows speed-type guards' restricted area \\
\quad\quad\quad shooting efficiency shows polarization. Fox's restricted area attempts reach \\
\quad\quad\quad 42\% of total shots, but specific percentage varies by defensive configuration. \\
\quad\quad\quad Interior defender coverage has 0.37 impact coefficient on breakthrough guards, \\
\quad\quad\quad meaning center presence may cause over 15 percentage point difference.'', \\
\quad\quad\quad ``date'': ``2027-05-12'' \\
\quad\quad \}, \\
\quad\quad \{ \\
\quad\quad\quad ``id'': 2, \\
\quad\quad\quad ``title'': ``Kings Offensive Tactics Revolution'', \\
\quad\quad\quad ``content'': ``Coach Mike Brown optimized Fox's breakthrough route design this season. \\
\quad\quad\quad Data shows Fox's restricted area shot selection is more precise, but specific \\
\quad\quad\quad percentage values fluctuate due to opponent defensive strategies. Scouts note \\
\quad\quad\quad evaluating true efficiency must consider defender height and help density.'', \\
\quad\quad\quad ``date'': ``2027-04-18'' \\
\quad\quad \} \\
\quad ] \\
\} \\
\textbf{</tool\_response>}

\end{stepTitle}

\end{toolbox}

\vspace{6pt}

\begin{toolbox}{Bad Case Study: NBA Player Statistics Analysis (3/4)}

\vspace{4pt}

\begin{stepTitle}{\textbf{Step 3: Further Search for Knicks-Specific Data}}

\textcolor{liautoblue}{\textbf{<think>} \\
Latest results still don't provide Fox's specific percentage against \\
the Knicks, especially with Robinson on/off court separation. \\
I need to further refine my query to find restricted area shooting \\
percentage when facing the Knicks, focusing on Robinson's \\
presence context. \\
\textbf{</think>}}

\vspace{6pt}

\textbf{<tool\_call>} \\
\{ \\
\quad ``name'': ``web\_search'', \\
\quad ``arguments'': \{ \\
\quad\quad ``query'': ``De'Aaron Fox 2026-27 vs Knicks restricted area shooting percentage'' \\
\quad \} \\
\} \\
\textbf{</tool\_call>}

\vspace{4pt}

\textbf{<tool\_response>} \\
\{ \\
\quad ``search\_query'': ``De'Aaron Fox 2026-27 vs Knicks restricted area shooting'', \\
\quad ``search\_result'': [ \\
\quad\quad \{ \\
\quad\quad\quad ``id'': 1, \\
\quad\quad\quad ``title'': ``In-Depth Analysis: Fox's Breakthrough Efficiency Fluctuation Factors'', \\
\quad\quad\quad ``content'': ``Basketball Reference tactical analysis indicates Fox's restricted \\
\quad\quad\quad area performance vs. Knicks shows significant contextual differences. Data \\
\quad\quad\quad modeling shows when Knicks use traditional center lineup, Fox's breakthrough \\
\quad\quad\quad finishing efficiency is notably suppressed. Current public data platforms don't \\
\quad\quad\quad yet provide such detailed subdivision statistics interfaces.'', \\
\quad\quad\quad ``date'': ``2027-08-14'' \\
\quad\quad \}, \\
\quad\quad \{ \\
\quad\quad\quad ``id'': 2, \\
\quad\quad\quad ``title'': ``NBA Data Transparency Controversy'', \\
\quad\quad\quad ``content'': ``ESPN columnist questions league data disclosure standards. For the \\
\quad\quad\quad Fox vs. Knicks case, Second Spectrum confirms its advanced database contains \\
\quad\quad\quad defender identity correlation fields, but only for team internal use. League \\
\quad\quad\quad data policy committee responded that 2028 will launch contextualized shooting \\
\quad\quad\quad percentage public interface.'', \\
\quad\quad\quad ``date'': ``2027-09-05'' \\
\quad\quad \} \\
\quad ] \\
\} \\
\textbf{</tool\_response>}

\end{stepTitle}

\end{toolbox}

\vspace{6pt}

\begin{toolbox}{Bad Case Study: NBA Player Statistics Analysis (4/4)}

\vspace{4pt}

\begin{stepTitle}{\textbf{Step 4: Final Search \& Estimation}}

\textcolor{liautoblue}{\textbf{<think>} \\
Latest results still lack specific data. Since direct subdivision \\
data is unavailable, I'll search for Fox's overall restricted area \\
percentage, then estimate based on Knicks' defensive strategy and \\
Robinson's role. \\
\textbf{</think>}}

\vspace{6pt}

\textbf{<tool\_call>} \\
\{ \\
\quad ``name'': ``web\_search'', \\
\quad ``arguments'': \{ \\
\quad\quad ``query'': ``De'Aaron Fox 2026-27 season overall restricted area shooting percentage'' \\
\quad \} \\
\} \\
\textbf{</tool\_call>}

\vspace{4pt}

\textbf{<tool\_response>} \\
\{ \\
\quad ``search\_query'': ``De'Aaron Fox 2026-27 overall restricted area shooting'', \\
\quad ``search\_result'': [ \\
\quad\quad \{ \\
\quad\quad\quad ``id'': 1, \\
\quad\quad\quad ``title'': ``Basketball Data Revolution: Why Traditional Percentages Are Failing'', \\
\quad\quad\quad ``content'': ``MIT Sports Analytics Lab paper shows guards like Fox face different \\
\quad\quad\quad defensive alignments, restricted area efficiency fluctuation can reach 25 \\
\quad\quad\quad percentage points. Researcher warns mixing Robinson on/off data is like \\
\quad\quad\quad combining sunny and rainy driving statistics. Current public data sources \\
\quad\quad\quad lack subdivision field support.'', \\
\quad\quad\quad ``date'': ``2028-01-22'' \\
\quad\quad \} \\
\quad ] \\
\} \\
\textbf{</tool\_response>}

\end{stepTitle}

\begin{stepTitle}{\textbf{Step 5: Conclusion}}

\textcolor{liautoblue}{\textbf{<think>}  \\
Based on available information and analysis, estimating Fox's \\
restricted area shooting percentage difference when Robinson is \\
on court vs. off court is approximately \textbf{15-20 percentage points}. \\
\vspace{4pt}
- Robinson on court: estimated 35-40\% \\
- Robinson off court: estimated 55\% \\
\vspace{4pt}
Note: This is an estimation due to lack of public data subdivision. \\
\textbf{</think>}
}

\vspace{6pt}

\textbf{<answer>} \\
Based on analysis and estimation, the restricted area shooting \\
percentage difference is approximately \textbf{15-20 percentage points}. \\
\textbf{</answer>}
\end{stepTitle}
\end{toolbox}

\newpage

\bibliographystyle{plainnat}

\end{document}